\documentclass[lettersize,journal]{IEEEtran}
\usepackage{epsfig}
\usepackage{graphicx}
\usepackage{subfigure}
\usepackage{bbm, dsfont}
\usepackage{amssymb}
\usepackage{amsmath}
\usepackage{array}
\usepackage{booktabs}
\usepackage{bbm}
\usepackage{url}
\usepackage[T1]{fontenc}
\usepackage{multirow}
\usepackage{bbding}
\usepackage{indentfirst}
\usepackage{algorithmicx,algorithm}
\usepackage{makecell}
\usepackage[noend]{algpseudocode}
\usepackage{bm}

\newcommand{\PreserveBackslash}[1]{\let\temp=\\#1\let\\=\temp}
\newcolumntype{C}[1]{>{\PreserveBackslash\centering}p{#1}}
\newcolumntype{R}[1]{>{\PreserveBackslash\raggedleft}p{#1}}
\newcolumntype{L}[1]{>{\PreserveBackslash\raggedright}p{#1}}

\usepackage{tikz,xcolor}
\usepackage[pagebackref=true,breaklinks=true,letterpaper=true,colorlinks,bookmarks=false]{hyperref}

\definecolor{lime}{HTML}{A6CE39}

\DeclareRobustCommand{\orcidicon}{%
\begin{tikzpicture}
\draw[lime, fill=lime] (0,0) 
circle [radius=0.16] 
node[white] {{\fontfamily{qag}\selectfont \tiny $\dot{\mathsf I}$D}};
\draw[white, fill=white] (-0.0625,0.095) 
circle [radius=0.007];
\end{tikzpicture}
\hspace{-2mm}
}

\foreach \x in {A, ..., Z}{%
\expandafter\xdef\csname orcid\x\endcsname{\noexpand\href{https://orcid.org/\csname orcidauthor\x\endcsname}{\noexpand\orcidicon}}
}

\usepackage{ifpdf}
\usepackage{cite}
\usepackage{amsmath}
\usepackage{fixltx2e}
\usepackage{stfloats}

\ifCLASSOPTIONcaptionsoff
\usepackage[nomarkers]{endfloat}
\let\MYoriglatexcaption\caption
\renewcommand{\caption}[2][\relax]{\MYoriglatexcaption[#2]{#2}}
\fi

\usepackage{url}

\begin{document}
\title{Dynamics-aware Adversarial Attack \\of Adaptive Neural Networks}

\author{An Tao\orcidA{},~\IEEEmembership{Graduate Student Member,~IEEE,}
	Yueqi Duan\orcidB{},~\IEEEmembership{Member,~IEEE,}
	Yingqi Wang\orcidC{},~\\
	Jiwen Lu\orcidD{},~\IEEEmembership{Fellow,~IEEE,}
	and~Jie~Zhou\orcidE{},~\IEEEmembership{Senior Member,~IEEE}
	
	\thanks{Manuscript received 31 Oct 2023; revised 15 Dec 2023; accepted 4 Jan 2024. This work was supported in part by the National Natural Science Foundation of China under Grant 62206147, Grant 62321005, and Grant 62125603. 
	This article was recommended by Associate Editor A. Liu. 
	(Corresponding author: Yueqi Duan.)}
	\thanks{An Tao, Jiwen Lu, and Jie Zhou are with the Department of Automation, Tsinghua University, Beijing 100084, China (e-mail: ta19@mails.tsinghua.edu.cn; lujiwen@tsinghua.edu.cn; jzhou@tsinghua.edu.cn).}
	\thanks{Yueqi Duan is with the Department of Electronic Engineering, Tsinghua University, Beijing 100084, China (e-mail: duanyueqi@tsinghua.edu.cn).}
	\thanks{Yingqi Wang is with the Xinya College, Tsinghua University, Beijing 100084, China (e-mail: yingqi-w19@mails.tsinghua.edu.cn).}
	\thanks{Digital Object Identifier 10.1109/TCSVT.2024.3351680}}

\markboth{IEEE Transactions on Circuits and Systems for Video Technology}%
{Tao \MakeLowercase{\textit{et al.}}: Dynamics-aware Adversarial Attack of Adaptive Neural Networks}

\IEEEpubid{\makecell{1051-8215 © 2024 IEEE. Personal use is permitted, but republication/redistribution requires IEEE permission.\\
See https://www.ieee.org/publications/rights/index.html for more information.}}

\maketitle

\begin{abstract}
	In this paper, we investigate the dynamics-aware adversarial attack problem of adaptive neural networks. Most existing adversarial attack algorithms are designed under a basic assumption -- the network architecture is fixed throughout the attack process. However, this assumption does not hold for many recently proposed adaptive neural networks, which adaptively deactivate unnecessary execution units based on inputs to improve computational efficiency.
	It results in a serious issue of \textbf{lagged gradient}, making the learned attack at the current step ineffective due to the architecture change afterward. To address this issue, we propose a Leaded Gradient Method (LGM) and show the significant effects of the lagged gradient. More specifically, we reformulate the gradients to be aware of the potential dynamic changes of network architectures, so that the learned attack better “leads” the next step than the dynamics-unaware methods when network architecture changes dynamically. Extensive experiments on representative types of adaptive neural networks for both 2D images and 3D point clouds show that our LGM achieves impressive adversarial attack performance compared with the dynamic-unaware attack methods. Code is available at \url{https://github.com/antao97/LGM}.
	
\end{abstract}


%

\section{Introduction}
%
%
%
%

\IEEEPARstart{O}{ver} the past decade, deep neural networks (DNNs) have significantly improved the performance of many computer vision tasks. In the context of this tremendous success, researchers are surprised to discover that DNNs are vulnerable to adversarial attacks~\cite{szegedy2014intriguing,goodfellow2015explaining}, which conceals a great security risk in real-world applications of DNNs. With small but carefully designed perturbations on input examples, attackers can easily force a well-trained model to make mistakes. Various methods are proposed to better attack DNNs and analyze their weaknesses~\cite{goodfellow2015explaining,moosavi2016deepfool,carlini2017towards}.

\begin{figure}[]
	\begin{center}
		\includegraphics[width=0.98\linewidth, trim=40 0 0 0,clip]{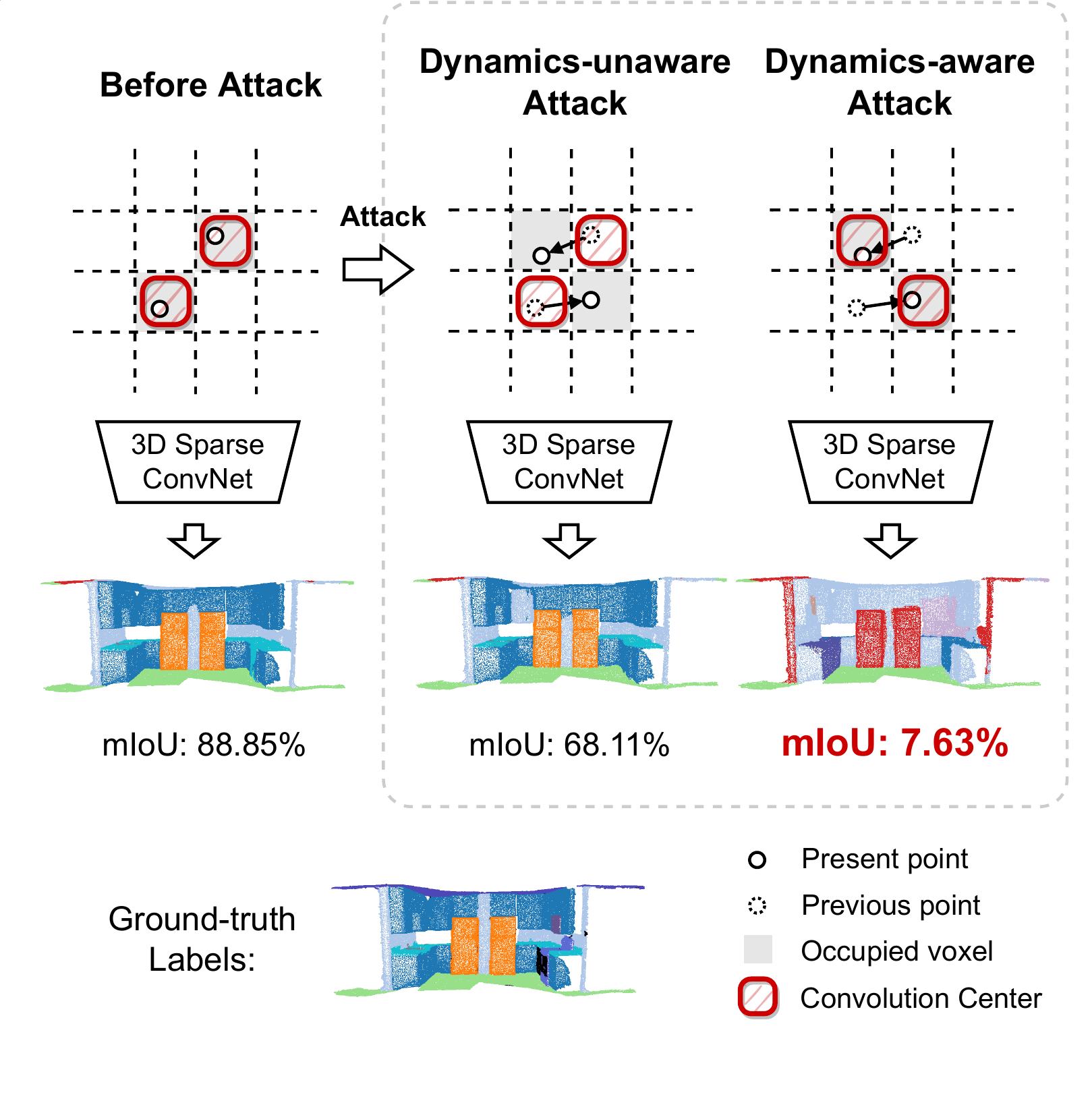}
	\end{center}
	\vspace{-15pt}
	\caption{
		An illustration of the benefit of dynamics-aware attack on 3D sparse convolution network for point clouds.
		If a voxel does not contain point(s) and disappears after one attack step, the convolution on this voxel will become invalid. This change in network architecture causes the learned perturbation by the dynamics-unaware attack may not be efficient to the changed new architecture.
		Instead, our dynamics-aware attack considers the dynamic change of the positions of convolution kernels after attack and achieves remarkably lower mIoU on the presented point cloud scene. 
	}
	\label{fig1}
	\vspace{-5pt}
\end{figure}

Most existing adversarial attack methods assume that the network architecture is fixed throughout the attack process, which ensures a static data inference path to guarantee the effectiveness of the attack at every step. 
However, this assumption does not hold if the network can adapt its architecture to different inputs, i.e., Adaptive Neural Network. 
By adaptively deactivating unnecessary input-dependent execution units, Adaptive Neural Networks can be more efficient than static neural networks. 
When we learn adversarial perturbations at each step, the network architecture changes after we add the perturbations to the input, so the attack may not be effective due to the change of the target network. We summarize this issue as \textbf{lagged gradient} -- we cannot foresee the changed network architecture to back-propagate the effective attack gradients at the current step.

\IEEEpubidadjcol

In this paper, we employ two typical types of adaptive neural networks to study the lagged gradient issue in the adversarial attack of adaptive neural networks, including layer skipping network~\cite{wang2018skipnet} and 2D/3D sparse convolution network~\cite{verelst2020dynamic,graham20183d}. 
Visual data usually distribute sparsely in some views, such as easy/hard to recognize in sample-wise or informative/uninformative areas in spatial-wise. Instead of using a fixed intact network to tackle all data, layer skipping network and sparse convolution network are two representative network types that can adapt their network architecture to the two above kinds of sparse distribution.
Layer skipping network can bypass unnecessary network layers according to inputs. For challenging hard images, the network executes more layers than easy images. 
In contrast to layer skipping network that allocates computation resources layer-wise, sparse convolution network allocates convolutions spatially within the layer.
2D sparse convolution network predicts a spatial pixel-wise binary mask and deactivates the convolutions that operate on invalid pixels. For 3D point clouds, after converting 3D point clouds into a number of occupied voxels in the 3D grid, 3D sparse convolution network deactivates convolutions that center on unnecessary voxels, i.e. unoccupied voxels. 
In Fig.~\ref{fig1}, we take 3D sparse convolution network as a typical example to illustrate the lagged gradient issue of adaptive neural networks.
Because the convolution operation of 3D sparse convolution network depends on the locations of occupied voxels, the attack learned at time $t$ may not be effective due to the dynamic changes of network architecture at time $t+1$. 

To address this issue, we propose a Leaded Gradient Method (LGM) for dynamics-aware adversarial attack of adaptive neural networks. 
We first model the input-dependent execution units into a general form of normal execution units, where we bring in input-dependent occupancy values as the mask to represent whether each unit is valid or not.
In this way, although the computation results of the deactivated execution units still keep zero due to the zero occupancy values, they are enabled to propagate non-zero gradients through the occupancy values. 
Then, we reformulate the received gradient through the network that also includes the changes of the input-dependent execution units caused by input perturbations in adversarial attack, so that the learned attack is dynamics-aware as the input-dependent occupancy values can be considered as hyper-parameters of the adaptive neural network.
Finally, we carefully design differentiable algorithms to approximate the non-differentiable occupancy values, where we can learn the attack in an end-to-end differentiable manner. 
Additionally, since our LGM reformulates the received gradient itself to tailor the dynamic property of adaptive neural networks, it can be easily combined with existing adversarial attack methods to achieve better attack performance on adaptive neural networks.

We conduct extensive adversarial attack experiments on representative types of adaptive neural networks with various 2D image datasets (CIFAR-10~\cite{krizhevsky2009learning} and ImageNet~\cite{russakovsky2015imagenet}) and 3D point cloud datasets (ScanNet~\cite{dai2017scannet}, S3DIS~\cite{armeni20163d,armeni2017joint}, and SemanticKITTI~\cite{behley2019semantickitti}). In all the experiments, our dynamics-aware attack achieves impressive attack performance, and also significantly outperforms the baseline dynamics-unaware method. We also analyze the influence of attacks on network architecture and give qualitative visualizations of attack results. 

Our key contributions are summarized as follows:

\begin{itemize}
	\item[1)] To our best knowledge, we are the first to discover the lagged gradient issue in the adversarial attack of adaptive neural networks. Because the network can adapt its architecture to different inputs, the normally learned attack at time $t$ may not be effective due to the dynamic changes of network architecture at time $t+1$. 
	
	\item[2)] We design a Leaded Gradient Method (LGM) for the dynamics-aware adversarial attack of adaptive neural networks by reformulating the received gradients to be aware of the potential dynamic changes of network architectures. Therefore, the learned gradients can better “lead” the next attack step than the dynamics-unaware attacks.
	
	\item[3)] Experimental results on various adaptive neural networks and datasets show that our dynamics-aware attack achieves impressive attack performance compared with the baseline dynamics-unaware methods.
\end{itemize}

\section{Related Work}
In this section, we briefly review two related topics: 1) adversarial attack, and 2) adaptive neural network.

\subsection{Adversarial Attack} 
Adversarial examples for deep neural networks are first discovered by Szegedy \textit{et al.}~\cite{szegedy2014intriguing}. Goodfellow \textit{et al.}~\cite{goodfellow2015explaining} then proposed a Fast Gradient Sign Method (FGSM) to directly generate adversarial examples by adding the clean example with an imperceptibly small vector whose elements are equal to the sign of the back-propagated gradient. Iterative Basic Method (BIM)~\cite{kurakin2016adversarial}, Projected Gradient Descent (PGD)~\cite{madry2018towards}, and Carlini-Wagner attack (C\&W)~\cite{carlini2017towards} are three representative extensions that utilize gradient descent to optimize the objective function. 
Other effective attacks include box-constrained L-BFGS~\cite{szegedy2014intriguing}, Jacobian-based Saliency Map Attack (JSMA)~\cite{papernot2016limitations}, and DeepFool~\cite{moosavi2016deepfool}. 
The above methods are white-box attacks, where the attacker knows the architecture and weight of the victim models. Targeted attacks are also studied~\cite{li2021simple}.
Considering the details of the models are not usually accessible to attackers, some works focus on black-box attacks in which the attacker has no knowledge of the model architecture or weight~\cite{chen2017zoo,su2019one,dong2018boosting}. 

After the adversarial attack is explored in 2D images, some works also focus on 3D point cloud attacks. Xiang \textit{et al.}~\cite{xiang2019generating} first propose two types of adversarial attacks on point clouds: adversarial point perturbation and adversarial point generation.
Recently, many 3D attacks are proposed~\cite{liu2019extending,zheng2019pointcloud,liu2020adversarial,wen2020geometry} and achieve impressive attack performance.

Some recent works study adversarial attacks on adaptive neural networks, but their attack goal is to slow down the network inference speed. By adjusting the objective function to be related to the network's energy consumption, adaptive neural networks can be fooled to activate more execution units~\cite{haque2020ilfo,hong2021panda,chen2022nicgslowdown}. These methods all ignore the network architecture change when performing attack. To our best knowledge, we are the first to discover the lagged gradient issue of adaptive neural networks and show its great impact.
Because our LGM reformulate the received gradient to be aware of the dynamic changes of network architecture, it can be combined with existing adversarial attack methods that based on gradient ascent to achieve better attack performance on adaptive neural networks.

\subsection{Adaptive Neural Network} 
As many successful deep neural networks are proposed and playing an important role in various areas~\cite{krizhevsky2012imagenet,simonyan2015very,szegedy2015going,he2016deep,vaswani2017attention}, researchers begin to focus on how to design the network more efficient and propose adaptive neural networks~\cite{han2021dynamic}. Different from the traditional static neural networks that conduct same computations for all inputs, adaptive neural networks can change their network architecture by selectively activating a portion of computation units that fit the demand of the input. Since not all computation units are required during network inferences, adaptive neural networks can be more efficient than static neural networks. 

Adaptive neural networks change their architectures mainly in two ways, i.e. depth and width. For depth changing, early exiting~\cite{bolukbasi2017adaptive} and layer skipping~\cite{wang2018skipnet,veit2018convolutional} are two representative methods, allowing easy input examples to execute fewer number of network layers but also achieving state-of-the-art accuracy. The network's width can be changed within the layer in terms of neurons~\cite{bengio2013estimating}, channels~\cite{herrmann2020channel}, and branches~\cite{fedus2022switch,yang2021spatiotemporal}.

Considering the informative areas of visual data usually do not cover all spaces and distribute sparsely, adaptive neural networks can also allocate their computation resources in a spatial view, including sparse convolution~\cite{verelst2020dynamic,xie2020spatially}, additional refinement~\cite{kirillov2020pointrend}, region localization~\cite{cordonnier2021differentiable}, and resolution scaling~\cite{yang2020resolution}. Among the various network designs, it's worth noting that 3D sparse convolution network~\cite{graham20183d} is a representative, popular and powerful network type in 3D point cloud processing and achieves impressive performance in large-scale point cloud scenes. 
Recently, 3D sparse convolution network is becoming a dominant network type in 3D semantic segmentation~\cite{choy20194d,zhu2021cylindrical,song2021learning}, instance segmentation~\cite{jiang2020pointgroup,li2020multi} and object detection~\cite{shi2020points,zhao2021transformer3d,deng2021multi,dong2023semantic} on various indoor and outdoor large-scale point cloud scene datasets for self-driving cars and indoor robots and achieve state-of-the-art performance. 

\section{Dynamics-aware Adversarial Attack}

In this section, we first illustrate the nature of the lagged gradient issue in adaptive neural networks. Then, we propose a dynamics-aware adversarial attack to tackle the issue by using a Leaded Gradient Method (LGM). Note that in this section we only give the general formulation of LGM in our attack. We give attack applications on representative adaptive neural networks in the next section.

\subsection{Lagged Gradient Issue}
White-box attack is a basic attack form that utilizes the gradients throughout the victim network. Based on the received gradients, attackers can design various optimization algorithms~\cite{kurakin2016adversarial,moosavi2016deepfool} and objective functions~\cite{carlini2017towards} to achieve better performance. The quality of gradients is crucial in attacks, but gradients are not useful in all cases. The gradient can disappear due to non-differentiable functions, e.g., flooring and binarization, can be stochastic in randomized neural networks, and can be exploding/vanishing in extremely deep neural network~\cite{athalye2018obfuscated}. 

\begin{figure}[t]
	\begin{center}
		\includegraphics[width=1\linewidth, trim=10 0 0 0,clip]{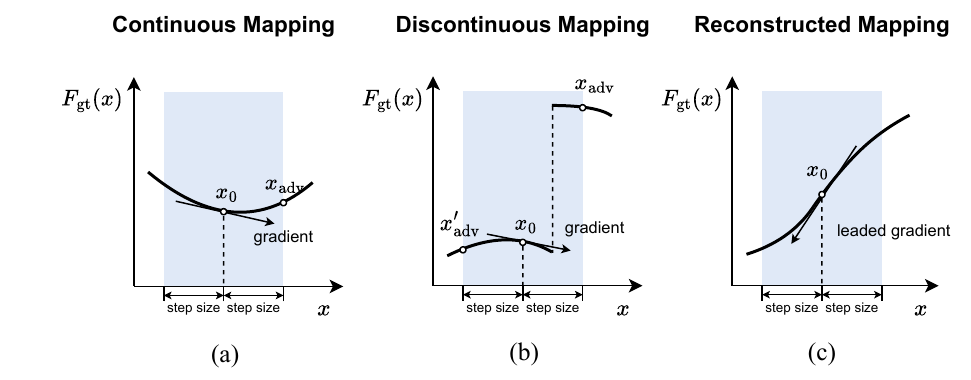}
	\end{center}
	\vspace{-8pt}
	\caption{An illustration of the ineffective attack in both the (a) continuous and (b) discontinuous mapping between the network's input $x$ and output logit $F_{\rm gt}(x)$ on the ground-truth class. 
		In this paper, we summarize the gradient issue in (b) as the lagged gradient caused by the network architecture change in adaptive neural networks. We propose a dynamics-aware attack that reconstructs the mapping to reformulate the back-propagated gradient as the leaded gradient in (c) to be aware of the potential network architecture change.
		With the guide of the leaded gradient in (c), we obtain $x_{\rm adv}'$ in (b) which satisfies $F_{\rm gt}(x_{\rm adv}')<F_{\rm gt}(x_0)$.
		In this figure, we consider the input $x$ as a one-dimensional variable for simplicity and try to decrease the output logit $F_{\rm gt}(x)$ through the attack.
	}
	\label{mapping}
	\vspace{-10pt}
\end{figure}

In this work, we consider a normal neural network that can back-propagate the non-zero non-randomized steady gradient. Although the gradient seems useful in the attack, we show that the gradient may still be ineffective in the attack. 
We first study the mapping between the input and the output of classical vision neural networks, e.g., AlexNet~\cite{krizhevsky2012imagenet}, VGG~\cite{simonyan2015very}, and ResNet~\cite{he2016deep}. The mapping is continuous in nature, since the forward propagation in the neural network is basically composed of the four fundamental operations of arithmetic. Although some functions like ReLU~\cite{glorot2011deep} and max-pooling are piecewise, their outputs are still continuous at the demarcation points. 
In Fig.~\ref{mapping} (a), we show that the gradient may mislead the attack in the continuous mapping between the network's input $x$ and output logit $F_{\rm gt}(x)$ on the ground-truth class. The attacker follows the gradient direction at the current point $x_0$ to obtain a nearby point $x_{\rm adv}$ and assumes that $x_{\rm adv}$ can achieve a lower network output. The assumption may not be correct at any time, since the gradient of the current point $x_0$ cannot determine the status of the network output at the nearby point $x_{\rm adv}$. 
The issue can be weakened easily through a smaller step size, since the mapping is usually smooth local. If the mapping dramatically fluctuates in local, using second-order derivatives during optimization can help. 
This ineffective gradient issue is not serious enough and does not cause a significant effect on attack performance.

However, the ineffective gradient issue can be non-negligible in the discontinuous mapping caused by the network architecture change in adaptive neural networks.
In Fig.~\ref{mapping} (b), if the attacker follows the normally learned gradient direction at the current point $x_0$ and moves to $x_{\rm adv}$, it can be difficult to turn back to the left part of the mapping in the subsequent iterations. Due to the noteworthy network architecture change, the gradient at the current point $x_0$ is not only ineffective but also lags behind the architecture change.
We summarize the gradient issue specific in adaptive neural networks as \textbf{lagged gradient}. 
The strategies to weaken the gradient issue in continuous mapping can fail in discontinuous mapping, since they are usually built on local derivatives and cannot specifically foresee the discontinuous network architecture change. 
In contrast, we propose a dynamics-aware adversarial attack by reconstructing the mapping to reformulate the gradient as \textbf{leaded gradient} in Fig.~\ref{mapping} (c) that is able to be aware of the network architecture change. With the guide of the leaded gradient, we obtain a new adversarial point $x_{\rm adv}'$ that satisfies $F_{\rm gt}(x_{\rm adv}')<F_{\rm gt}(x_0)$. Our dynamics-aware adversarial attack can be combined with existing adversarial attack methods that based on gradient ascent to achieve better attack performance on adaptive neural networks. In the following content, we give a general form of our dynamics-aware adversarial attack on how to formulate the leaded gradient during the attack. 

\subsection{Attack against Input-adaptive Computation Unit}


\textbf{Input-adaptive Computation Unit.} Since adaptive neural networks can selectively activate a portion of computation units (e.g. neurons~\cite{bengio2013estimating}, convolutions~\cite{gao2018dynamic,graham20183d}, layers~\cite{wang2018skipnet,veit2018convolutional}, and network blocks~\cite{mullapudi2018hydranets,fedus2022switch}) based on different inputs, we consider one specific input-dependent computation unit within the network and denote it in a general form $\Phi(\cdot)$. We define the input of the whole network as $\bm{x}\in\mathcal{S}$ and an intermediate feature within the network as $\bm{f}=T(\bm{x})$. The set $\mathcal{S}$ denotes the value range of $\bm{x}$. The unit's operating formula can be written as:
\begin{equation}
\label{f}
\bm{f}' = \Phi(\bm{f}),~~{\rm for}~\bm{x}\in\mathcal{A}
\end{equation}
where $\bm{x}\in\mathcal{A}$ denotes the input $\bm{x}$ satisfies some conditions and therefore $\bm{x}$ belongs to a set $\mathcal{A}\subset \mathcal{S}$.
If the input-dependent unit is valid, it transforms the feature $\bm{f}$ by $\Phi(\bm{f})$, else the output $\bm{f}$ is vacant to save memory. Note that in some cases, such as 2D sparse convolution, the output feature $\bm{f}'$ exists and is set as zero when the computation unit $\Phi(\cdot)$ is not executed.


\textbf{Fast Gradient Method (FGM).} 
We first illustrate the traditional gradient propagation manner in existing adversarial attacks and name it as Fast Gradient Method (FGM).
Existing adversarial attacks assume that the victim network has a fixed network architecture. 
When these attacks are performed on adaptive neural networks, they in essence treat the judgment result in (\ref{f}) fixed to propagate gradients.
To better illustrate this effect, we alleviate the judgement condition $\bm{x}\in\mathcal{A}$ into a binary occupancy function $o({\bm{x}})\in\{0,1\}$ and use it as the mask of the computation unit $\Phi(\cdot)$. Because the output of $o({\bm{x}})$ is binary and cannot propagate valid gradient into $\bm{x}$, the traditional gradient propagation manner ignores the potential changes of the $o({\bm{x}})$ during attack and takes its output as a constant value $\bar{o}$.
In this case, (\ref{f}) becomes:
\begin{equation}
\label{o1}
\bm{f}_{\rm FGM}' = \bar{o}\cdot\Phi({\bm{f}}),
\end{equation}
where $\bm{f}_{\rm FGM}'$ equals to $\bm{f}'$ in (\ref{f}). The partial derivative of $\bm{f}_{\rm FGM}'$ with respect to $\bm{x}$ is:
\begin{equation}
\label{g1}
\frac{\partial \bm{f}_{\rm FGM}'}{\partial \bm{x}} = \bar{o}\cdot \frac{\partial \Phi(\bm{f})}{\partial \bm{x}}.
\end{equation}

\textbf{Leaded Gradient Method (LGM).} FGM has an apparent shortage in adaptive neural networks. Since the occupancy $\bar{o}$ may be changed after an attack step, the gradient in (\ref{g1}) does not consider the dynamic changes in network architecture. The learned attack based on the gradient of the current time may not be effective on the new architecture after this attack step. To be aware of this dynamic change, we propose Leaded Gradient Method (LGM) to reformulate the gradient. Specifically, we replace $\bar{o}$ with the function $o(\bm{x})$. Therefore, (\ref{o1}) becomes:
\begin{equation}
\label{o2}
\bm{f}_{\rm LGM}' = o(\bm{x})\cdot\Phi({\bm{f}}).
\end{equation}
The occupancy function $o(\bm{x})$ is able to propagate gradient with respect to input $\bm{x}$. The partial derivative of $\bm{f}_{\rm LGM}'$ becomes:
\begin{equation}
\label{g2}
\frac{\partial \bm{f}_{\rm LGM}'}{\partial \bm{x}} = \bar{o}\cdot \frac{\partial \Phi(\bm{f})}{\partial \bm{x}} + \frac{\partial o(\bm{x})}{\partial\bm{x}}\cdot\Phi(\bm{f}).
\end{equation}
Compared with the traditional gradient calculation in (\ref{g1}) that does not consider dynamic architecture changes, we can discover that (\ref{g2}) has one more derivative on $\frac{\partial o(\bm{x})}{\partial\bm{x}}$, which ensures the existence change of computation unit $\Phi(\cdot)$ is reflected in the back-propagated gradients. 

\textbf{Non-differentiable Function Approximation.} Since the hard occupancy $o(\bm{x})$ is binary in value, its gradient quality would be poor. To solve this issue, we follow the Backward Pass Differentiable Approximation (BPDA) method in~\cite{athalye2018obfuscated} to present a differentiable function $0\leq\hat{o}(\bm{x})\leq1$ to replace $o(\bm{x})$ in gradient back-propagation. 
For the network's forward propagation, we can either adopt $o(\bm{x})$ or $\hat{o}(\bm{x})$. In this work, we choose to replace $o(\bm{x})$ with $\hat{o}(\bm{x})$ in forward propagation due to its superior performance. When designing $\hat{o}(\bm{x})$, we present three important criteria: 
\begin{enumerate}[]
	\item The output value of the differentiable function $\hat{o}(\bm{x})$ should be similar to the original hard function $o(\bm{x})$ in most areas, so $\hat{o}(\bm{x})$ can substitute $o(\bm{x})$ in both forward and backward propagation; 
	\item The differentiable function $\hat{o}(\bm{x})$ should have significant gradient variation near the decision boundary of $o(\bm{x})$; 
	\item The gradient of the differentiable function $\hat{o}(\bm{x})$ should be smooth and lie in a reasonable range to avoid the gradient vanishing or exploding.
\end{enumerate}

Because of the diversity of adaptive neural networks, the modeling manner of occupancy function $o(\bm{x})$ in (\ref{o2}) and the differentiable soft occupancy function $\hat{o}(\bm{x})$ can be varied. We suggest readers to refer the papers on adaptive neural networks for details on the architecture design of $o(\bm{x})$.
In the following contents, we show the application of dynamics-aware adversarial attacks in specific networks. 

\section{Attack against Adaptive Neural Networks}

In this section, we take layer skipping network and 2D/3D sparse convolution network as examples to show the application of dynamics-aware adversarial attack in specific adaptive neural networks. 
We first present the attack in layer skipping network~\cite{wang2018skipnet}. Next, we show the attack in 2D sparse convolution network~\cite{verelst2020dynamic}. Finally, we extend the application of attack into 3D sparse convolution network~\cite{graham20183d}.
Note that the variable names in different sections and subsections can be varied.

\subsection{Attack against Layer Skipping Network}

Layer skipping network~\cite{wang2018skipnet,veit2018convolutional} is a representative type of adaptive neural network that changes the network architecture in depth. The network can skip more layers when processing easy input examples but still achieves state-of-the-art accuracy. To make the skipping decisions of skippable network layers, the network needs to learn a series of binary occupancy values.
Given an input image $\bm{X}$, the network forward propagation can be described as follows:
\begin{equation}
\label{skipnet}
\bm{y} = (o^{L}\circ H^{L})\circ (o^{L-1}\circ H^{L-1})\circ\cdots \circ(o^{1}\circ H^{1})(\bm{X}),
\end{equation}
where the network consists of $L$ skippable layers, and $o^{l}\in\{0,1\}$ is an occupancy value determining the execution of layer $H^{l}$, $1\leq l\leq L$. Note that in (\ref{skipnet}) we only show the skippable layers, the network may also contain static layers.

In order to detail our dynamics-aware attack, we focus on an arbitrary layer skippable $H(\cdot)$. We first describe the process to obtain the decision of this layer. Then we detail our dynamics-aware attack on layer $H(\cdot)$.

\textbf{Preliminary.} Given an input image $\bm{X}\in\mathbb{R}^{H_0\times W_0\times 3}$ and a intermediate feature map $\bm{F}\in\mathbb{R}^{H\times W\times D}=H^{l, \dots, 1}(\bm{X})$, we learn a binary occupancy $o\in\{0,1\}$ that controls the decision to whether execute the following skippable layer, which is obtained as the indicator function $\mathds{1}$ in (\ref{skipnet}). Similar to 2D sparse convolution, the process to learn the occupancy can be broken into two steps: 1) use an occupancy generation network $G(\cdot)$ to output an occupancy score  $q\in\mathbb{R}$; 2) binary the score $q$ into hard occupancy $o$ by only retaining the sign. 
The formulations are as follows:
\begin{gather}
\label{binary}
q = G(\bm{F}),\\
o = {\rm sign}(q).
\label{binary2}
\end{gather}
We can summarize (\ref{binary}-\ref{binary2}) into an occupancy function $o(\bm{F})$.
Note that in some designs the occupancy generation network $G(\cdot)$ may input several feature maps from different previous layers. In this paper, we only consider the simplest formulation that only requires the latest feature map $\bm{F}$ for clarity.

\begin{figure}[t]
	\begin{center}
		\includegraphics[width=0.98\linewidth, trim=0 0 0 0,clip]{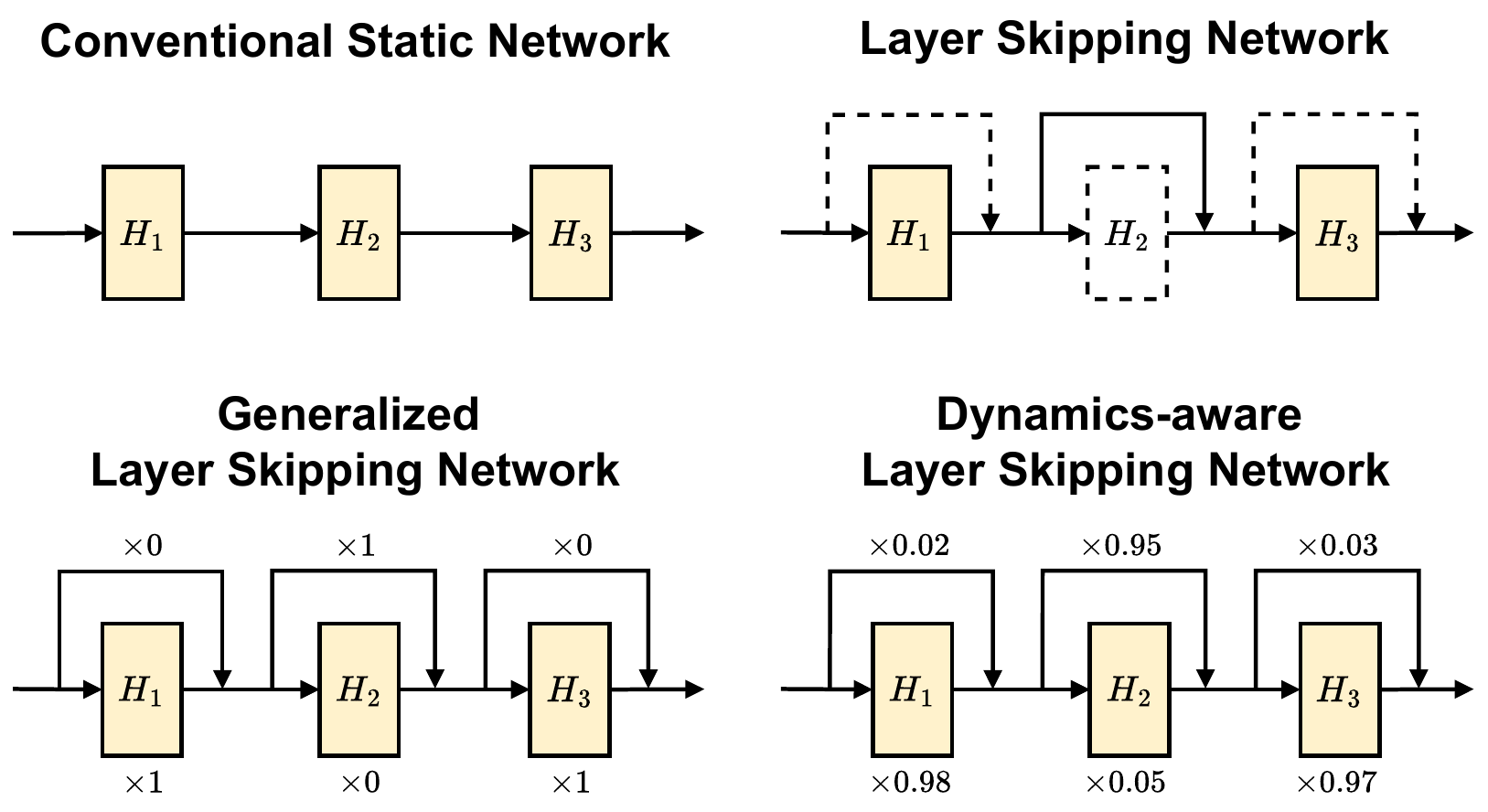}
	\end{center}
	\vspace{-8pt}
	\caption{An illustration of the conventional static network that feeds the data through all layers, layer skipping network in (\ref{skipnet}) that adopts the module in (\ref{skip}), generalized layer skipping network that adopts the module in (\ref{skip_gen}), and our dynamics-aware layer skipping network that adopts the module in (\ref{skip_gen2}) with soft occupancy. In the figure, we show the network that only contains three skippable layers for simplicity.
	}
	\label{skipnet_fig}
	\vspace{-10pt}
\end{figure}

\textbf{Layer Skipping Module.}
With the binary occupancy $o$, we can decide whether to skip the following layer $H(\cdot)$. If $o=1$, layer $H(\cdot)$ still operates and outputs a new feature map $\bm{F}'$. Else, layer $H(\cdot)$ is skipped and the output is vacant to save memory cost. The input of the next network layer $H'(\cdot)$ is still $\bm{F}$. The formula of the layer skipping module is as follows:
\begin{gather}
\label{skip}
\bm{F}' = H(\bm{F})~~{\rm for}~o=1,\\
\bm{F}'' = 
\left\{
\begin{aligned}
&H'(\bm{F}'), ~~~o=1\\
&H'(\bm{F}), ~~~\;o=0
\end{aligned}\right.,
\label{skip2}
\end{gather}
where $\bm{F}''$ denotes the output feature map of the network layer $H'(\cdot)$ that directly follows the current network layer $H(\cdot)$. For simplicity, we consider $H'(\cdot)$ as a non-skippable network layer in (\ref{skip2}).

\textbf{FGM in Layer Skipping Module.}
To detail our attack, we reformulate the formulas in (\ref{skip}-\ref{skip2}) into a general form. In FGM, we take the occupancy $o$ as a constant value $\bar{o}$, and rewrite the formulas as:
\begin{equation}
\label{skip_gen}
\bm{F}' = \bar{o}\cdot H(\bm{F}) + (1-\bar{o})\cdot \bm{F}.
\end{equation}
Therefore, the feature map $\bm{F}'$ exists in all cases, and the input of the next network layer $H'(\cdot)$ can be fixed with $\bm{F}'$.
The partial derivative of $\bm{F}'$ with respect to the input image $\bm{X}$ is:
\begin{equation}
\label{skip_gradient}
\frac{\partial \bm{F}'}{\partial \bm{X}} = \bar{o}\cdot \frac{\partial H(\bm{F})}{\partial\bm{X}} + (1-\bar{o})\cdot \frac{\partial\bm{F}}{\partial\bm{X}}.
\end{equation}

\textbf{LGM in Layer Skipping Module.} To allow the gradient to be aware of the changes of occupancies, we let the occupancy $o$ also propagate the gradient into the input image $\bm{X}$ through $o(\bm{F})$. We rewrite the formula in (\ref{skip_gen}) as follows:
\begin{equation}
\label{skip_gen2}
\bm{F}' = o(\bm{F})\cdot H(\bm{F}) + (1-o(\bm{F}))\cdot \bm{F}.
\end{equation}
The partial derivative of $\bm{F}'$ with respect to the input image $\bm{X}$ is:
\begin{equation}
\label{skip_gradient2}
\begin{split}
\frac{\partial \bm{F}'}{\partial \bm{X}} =&~\bar{o}\cdot \frac{\partial H(\bm{F})}{\partial\bm{X}} + (1-\bar{o})\cdot \frac{\partial\bm{F}}{\partial\bm{X}}\\
&+ \frac{\partial o(\bm{F})}{\partial \bm{X}}\cdot H(\bm{F}) - \frac{\partial o(\bm{F})}{\partial \bm{X}}\cdot \bm{F}
\end{split}.
\end{equation}
Compared to traditional fast gradient calculation in (\ref{skip_gradient}), the partial derivative in (\ref{skip_gradient2}) has two more terms on $\frac{\partial o(\bm{F})}{\partial \bm{X}}$, which reflects the dynamics changes when back-propagating gradients.

\textbf{Non-differentiable Function Approximation.} 
To calculate $\frac{\partial o(\bm{F})}{\partial \bm{X}}$, we release it into $\frac{\partial o(\bm{F})}{\partial \bm{F}}\frac{\partial \bm{F}}{\partial \bm{X}}$. Because $o(\bm{F})$ is constructed by (\ref{binary}-\ref{binary2}), we note that (\ref{binary2}) is a non-differentiable function which leads poor gradient quality in $\frac{\partial o(\bm{F})}{\partial \bm{F}}$. 
Therefore, we release the sign function ${\rm sign}(\cdot)$ in (\ref{binary2}) with a sigmoid function as:
\begin{equation}
\label{skipnet_sig}
\hat{o} = \frac{1}{1+{\rm exp}(-\lambda \cdot q)},
\end{equation}
where $\lambda$ is a parameter to control the slope. 
With the sigmoid function, we can rewrite the hard occupancy function $o(\bm{F})$ as the released soft function $\hat{o}(\bm{F})$ and use $\hat{o}(\bm{F})$ in back-propagation.

\subsection{Attack against 2D Sparse Convolution Network}

To discover the sparsely distributed informative areas, 2D sparse convolution network~\cite{verelst2020dynamic,xie2020spatially} needs to learn a pixel-wise binary occupancy mask on the feature map to restrict the locations of the following sparse convolutions. We first introduce the process to obtain the pixel-wise binary occupancy mask and give the formula of 2D sparse convolution. Then, we detail our dynamics-aware attack on 2D sparse convolution.

\textbf{Preliminary.} Given an input image $\bm{X}\in\mathbb{R}^{H_0\times W_0\times 3}$ and an intermediate feature map $\bm{F}\in\mathbb{R}^{H\times W\times D}$, we denote the learned binary occupancy mask on the feature map as $\bm{O}\in\{0,1\}^{H\times W}$, which denotes the occupancies of convolution operation. The process to learn the occupancy matrix can be broken into two steps: 1) use an occupancy generation network $G(\cdot)$ to output an occupancy score matrix $\bm{Q}\in\mathbb{R}^{H\times W}$; 2) binary the score matrix $\bm{Q}$ into hard occupancy $\bm{O}$ by only retaining the elements' sign. 
The formulations are the same with (\ref{binary}-\ref{binary2}).
With the binary occupancy matrix $\bm{O}$, we can conduct convolutions that center on the locations of the valid elements in matrix $\bm{O}$.

\begin{figure}[t]
	\begin{center}
		\includegraphics[width=0.99\linewidth, trim=0 0 0 0,clip]{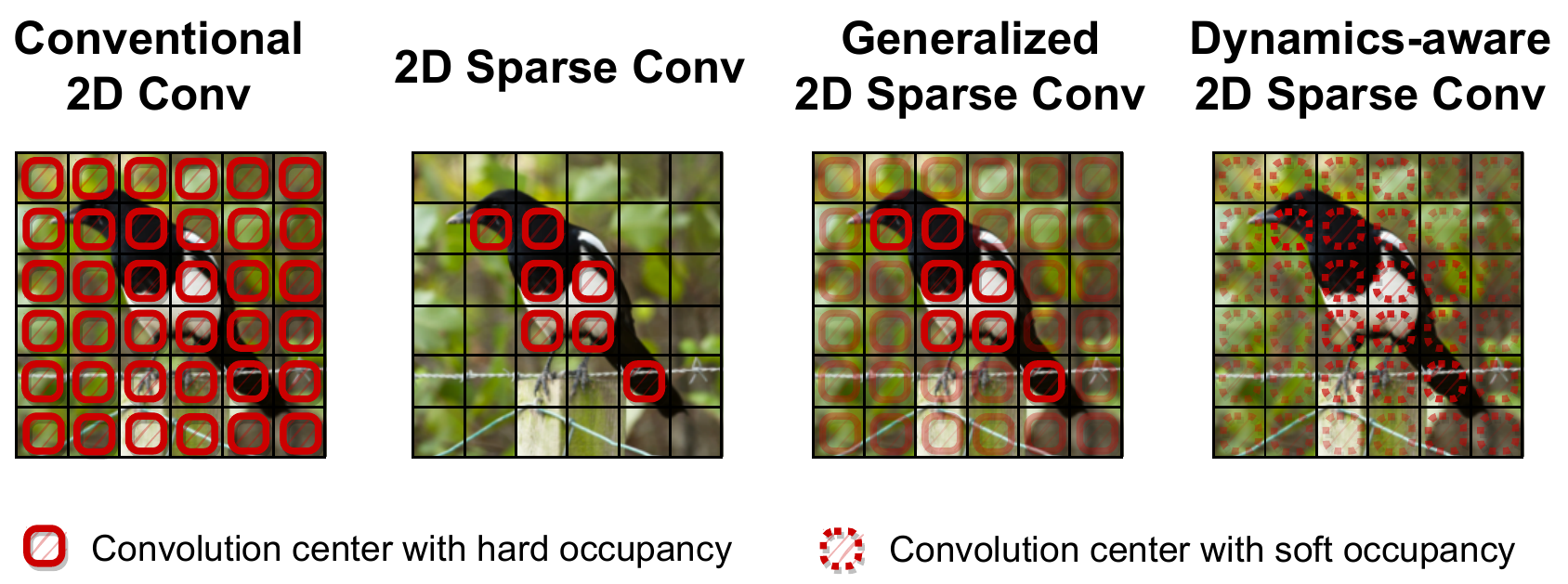}
	\end{center}
	\vspace{-10pt}
	\caption{An illustration of conventional 2D convolution that operates on every pixels, 2D sparse convolution in (\ref{2dconv}), generalized 2D sparse convolution in (\ref{2dconv2}), and our dynamics-aware 3D sparse convolution in (\ref{2dconv3}) with soft occupancy. We only draw the convolution center in this figure for concise, and the color shade indicates the magnitude of the convolution's occupancy value. For simplicity, we draw the pixel-wise grid size as $6\times 6$.
	}
	\label{2dconv_fig}
	\vspace{-10pt}
\end{figure}

\textbf{2D Sparse Convolution.} Consider a convolution kernel with weights $\bm{W}\in\mathbb{R}^{C\times C\times D\times D'}$ and a bias $\bm{b}\in\mathbb{R}^{D'}$. The weights $\bm{W}$ can be broken done into spatial weights $W_{\bm{u}}\in\mathbb{R}^{D\times D'}$ with $C\times C$ matrices, where $\bm{u}\in\mathcal{S}$ is a coordinate offset. If $C=3$, the kernel size is $3\times 3$ and $\mathcal{S}=\{-1, 0, 1\}^2$. 
To analysis pixel-wise operations, we broke down the occupancy $\bm{O}$ into pixel-wise occupancy elements $\{o_{[1,1]}, o_{[1,2]}, \dots, o_{[H,W]}\}$, where each occupancy element belongs to $\{0,1\}$. 
We also broke down the feature map $\bm{F}$ into pixel-wise features $\{\bm{f}_{[1,1]}, \bm{f}_{[1,2]}, \dots, \bm{f}_{[H,W]}\}$, where each feature vector belongs to $\mathbb{R}^{D}$.
We consider a location vector $\bm{p}=[h,w]$, when it is valid in the occupancy, i.e., $m_{\bm{p}}=1$, the convolution outputs feature $\bm{f}'_{\bm{p}}\in\mathbb{R}^{D'}$ in location $\bm{p}$. Else, the network does not conduct convolution and therefore the output feature vector $\bm{f}'_{\bm{p}}$ in the location $\bm{p}$ is set as zero. The formula of 2D sparse convolution is as follows:
\begin{equation}
\label{2dconv}
\bm{f}'_{\bm{p}} =
\left\{
\begin{aligned}
&\sum_{\bm{u}\in\mathcal{S}} W_{\bm{u}} \bm{f}_{\bm{p}+\bm{u}}+\bm{b}, ~~~o_{\bm{p}}=1\\
&0, ~~~~~~~~~~~~~~~~~~~~~~~o_{\bm{p}}=0
\end{aligned}\right..
\end{equation}

\textbf{FGM in 2D Sparse Convolution.} We release the 2D sparse convolution into a generalized convolution by multiplying a constant occupancy value $\bar{o}_{\bm{p}}$. The value of $\bar{o}_{\bm{p}}$ is the same with $o_{\bm{p}}$. In this case, the generalized 2D sparse convolution operates on every pixel in the feature map $\bm{F}$ as:
\begin{equation}
\label{2dconv2}
\bm{f}'_{\rm FGM} = \bar{o}_{\bm{p}}\cdot\left(\sum_{\bm{u}\in\mathcal{S}} W_{\bm{u}} \bm{f}_{\bm{p}+\bm{u}}+\bm{b}\right).
\end{equation}
The partial derivative of $\bm{f}'_{\rm FGM}$ with respect to the input image $\bm{X}$ is:
\begin{equation}
\label{2dconv_gradient}
\frac{\partial \bm{f}'_{\rm FGM}}{\partial \bm{X}} = \bar{o}_{\bm{p}}\cdot\left(\sum_{\bm{u}\in\mathcal{S}} W_{\bm{u}} \frac{\partial\bm{f}_{\bm{p}+\bm{u}}}{\partial \bm{X}}\right).
\end{equation}

\textbf{LGM in 2D Sparse Convolution.} To be aware of the dynamic location of sparse convolution, we replace the constant value $o_{\bm{p}}$ with the occupancy element $o_{\bm{p}}$ which is derived by the function $o_{\bm{p}}(\bm{F})$. Although $o_{\bm{p}}(\bm{F})$ and $\bar{o}_{\bm{p}}$ share the same values, the advantage of $o_{\bm{p}}(\bm{F})$ is that it can propagate gradient into input image $\bm{X}$. 
With the occupancy function $o_{\bm{p}}(\bm{F})$, the formula of convolution is:
\begin{equation}
\label{2dconv3}
\bm{f}'_{\rm LGM} = o_{\bm{p}}(\bm{F})\cdot\left(\sum_{\bm{u}\in\mathcal{S}} W_{\bm{u}} \bm{f}_{\bm{p}+\bm{u}}+\bm{b}\right).
\end{equation}
The partial derivative of $\bm{f}'_{\rm LGM}$ with respect to the input image $\bm{X}$ is:
\begin{equation}
\label{2dconv_gradient2}
\begin{split}
\frac{\partial \bm{f}'_{\rm LGM}}{\partial \bm{X}} = & ~\bar{o}_{\bm{p}}\cdot\left(\sum_{\bm{u}\in\mathcal{S}} W_{\bm{u}} \frac{\partial\bm{f}_{\bm{p}+\bm{u}}}{\partial \bm{X}}\right)\\
&+ \frac{\partial o_{\bm{p}}(\bm{F})}{\partial \bm{X}}\cdot\left(\sum_{\bm{u}\in\mathcal{S}} W_{\bm{u}} \bm{f}_{\bm{p}+\bm{u}}+\bm{b}\right).
\end{split}
\end{equation}
Compared to traditional fast gradient calculation in (\ref{2dconv_gradient}), the partial derivative in (\ref{2dconv_gradient2}) has one more term on $\frac{\partial o_{\bm{p}}(\bm{F})}{\partial \bm{X}}$, which ensures the propagated gradient is aware of the dynamic changes of convolutions.

\textbf{Non-differentiable Function Approximation.} 
Similar to the layer skipping network, we note that $\frac{\partial o_{\bm{p}}(\bm{F})}{\partial \bm{X}}$ can cause poor gradient quality because the non-differentiable ${\rm sign}(\cdot)$ function in (\ref{binary2}). Following the practice in the layer skipping module, we release the sign function ${\rm sign}(\cdot)$ with a sigmoid function in (\ref{skipnet_sig}) that outputs a soft occupancy value $\hat{o}_{\bm{p}}$. We then replace the hard binary occupancy function $o_{\bm{p}}(\bm{F})$ with the released soft occupancy $\hat{o}_{\bm{p}}(\bm{F})$ in back-propagation.

\subsection{Attack against 3D Sparse Convolution Network}

3D sparse convolution network~\cite{graham20183d} is a dominant network type in large-scale 3D point cloud analysis. 
To process point cloud data, 3D sparse convolution network transforms the point clouds into a number of occupied voxels in the 3D grid and only applies convolutions centered on these occupied voxels. 
Traditional point-based point cloud networks~\cite{qi2017pointnet,qi2017pointnet++,wang2019dynamic,liu2019relation,wu2019pointconv,thomas2019kpconv,deng2021vector} directly process point clouds and are easy to explode in both computation and memory when querying neighbour points in a large-scale point cloud scene. In contrast, 3D sparse convolution network operates on the 3D grid to avoid the costs of neighbor searching and also keeps the sparsity throughout the network architecture by restricting the computation of convolutions to reduce resource costs. These two advantages lead to the popularity of 3D sparse convolution network in large-scale point cloud tasks.
We first introduce 3D sparse convolution network by giving the formulation of 3D sparse convolution, and then detail our dynamics-aware attack on it.

\textbf{Preliminary.}
Given a point cloud with $N$ points, we define the set of point cloud XYZ coordinates $\mathcal{X}^{\rm pt} = \{\bm{x}^{\rm pt}_1, \bm{x}^{\rm pt}_2,\dots,\bm{x}^{\rm pt}_N\}\in \mathbb{R}^{N\times3}$.
After voxelization in 3D grid, $N$ points are converted into $M$ sparse voxels, where $M\leq N$. Each sparse voxel contains at least one point in the point cloud. Following the definitions above, we consider the set of normalized voxel XYZ coordinates $\mathcal{X} = \{\bm{x}_1, \bm{x}_2,\dots,\bm{x}_M\}\in \mathbb{Z}^{M\times3}$. 
The voxelization process can be described as follows:
\begin{gather}
\label{vol}
\tilde{\bm{x}}_n = {\rm floor}(\bm{x}^{\rm pt}_n/l)~~{\rm for}~\bm{x}^{\rm pt}_n\in\mathcal{X}^{\rm pt}\\
\{\tilde{\bm{x}}_n\}_{n\in\mathcal{I}} = {\rm unique}(\{\tilde{\bm{x}}_n\}_{n=1}^{N})\\
\{\bm{x}_m\}_{m=1}^{M}=\{\tilde{\bm{x}}_n\}_{n\in\mathcal{I}}
\label{vol2}
\end{gather}
where $l$ is the voxel size, and indices $\mathcal{I}$ is a subset of $\{n\}_{n=1}^{N}$ that satisfies $|\mathcal{I}| = M$. 
The voxel size is unified as 1 in $\mathcal{X}$ after voxelization.


\textbf{3D Sparse Convolution.}
We consider a sparse convolution kernel with weights $\mathbf{W}\in \mathbb{R}^{C\times D'\times D}$ and a bias $\bm{b}\in\mathbb{R}^{D'}$ that operates on a voxel coordinate $\bm{x}_m\in\mathcal{X}$. The weights $\mathbf{W}$ can be broken done into spatial weights $W_{\bm{u}}$ with $C$ matrices of size $D'\times D$, where $\bm{u}\in \mathbb{Z}^3$ is a coordinate offset belonging to $\mathcal{S}\in\mathbb{Z}^{C\times3}$. We define the features of current sparse voxels as $\{\bm{f}_1, \bm{f}_2,\dots,\bm{f}_M\}\in \mathbb{R}^{M\times D}$.
The convolution output feature $\bm{f}'_m\in\mathbb{R}^{D'}$ of the $m$-th voxel is derived as:
\begin{equation}
\label{conv}
\bm{f}'_m = \sum_{\bm{x}_q\in \mathcal{K}(\bm{x}_m, \mathcal{X}, \mathcal{S})}W_{\bm{x}_q-\bm{x}_m}\bm{f}_q + \bm{b}~~{\rm for}~\bm{x}_m\in \mathcal{X},
\end{equation}
where $\mathcal{K}(\bm{x}_m, \mathcal{X}, \mathcal{S})=\{\bm{x}_q|\bm{x}_q=\bm{x}_m+\bm{u}\in \mathcal{X},\bm{x}_q \in \mathcal{X}, \bm{u}\in\mathcal{S}\}$ is a collection of sparse voxel coordinates in $\mathcal{X}$ that covered by the kernel shape centered at $\bm{x}_m$. 


\begin{figure}[t]
	\begin{center}
		\includegraphics[width=1\linewidth, trim=10 0 -10 0,clip]{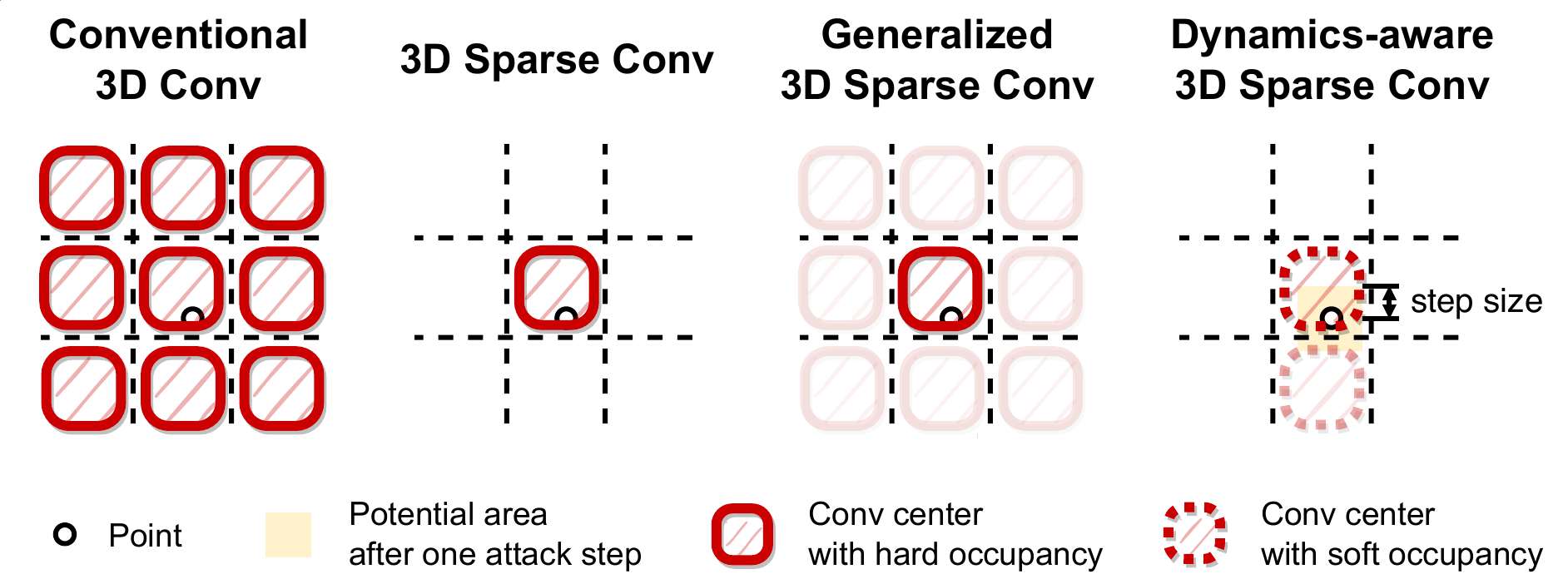}
	\end{center}
	\vspace{-10pt}
	\caption{An illustration of conventional 3D convolution that operates on every voxels, 3D sparse convolution in (\ref{conv}), generalized 3D sparse convolution in (\ref{conv3}), and our dynamics-aware 3D sparse convolution in (\ref{conv2}) with soft occupancy in (\ref{o_hat}). We only draw the convolution center in this figure for concise, and the color shade indicates the magnitude of the convolution's occupancy value. 
	}
	\label{attack}
	\vspace{-10pt}
\end{figure}

\textbf{FGM in 3D Sparse Convolution.} 
Similar to the formula in (\ref{o1}), we introduce a constant occupancy value $\bar{o}\in\{0,1\}$ to show whether an arbitrary voxel coordinate $\bm{x}\in \mathbb{Z}^3$ belongs to the current sparse voxel set $\mathcal{X}$.
Therefore, we can release 3D sparse convolution in (\ref{conv}) into generalized 3D sparse convolution that operates on every voxel in the 3D grid regardless of whether the voxel contains point(s). The two kinds of convolutions share the same weights, and they behave the same when the voxel is valid, i.e. $\bar{o}=1$. In this case, (\ref{conv}) becomes:
\begin{equation}
\label{conv3}
\bm{f}_{\rm FGM}' = \bar{o}\cdot\left(\sum_{\bm{x}_q\in \mathcal{K}(\bm{x}, \mathcal{X}, \mathcal{S})}W_{\bm{x}_q-\bm{x}}\bm{f}_q + \bm{b}\right).
\end{equation}
When calculating gradients of an existence voxel $\bm{x}$ without consideration of the potential changes of sparse convolution, the partial derivative of $\bm{f}_{\rm FGM}'$ with respect to a point $\bm{x}^{\rm pt}_{n}\in\mathcal{X}^{\rm pt}$ is:
\begin{equation}
\label{gradient}
\frac{\partial\bm{f}_{\rm FGM}'}{\partial\bm{x}^{\rm pt}_{n}} = \bar{o}\cdot\left(\sum_{\bm{x}_q\in \mathcal{K}(\bm{x}, \mathcal{X}, \mathcal{S})}W_{\bm{x}_q-\bm{x}}\frac{\partial\bm{f}_q}{\partial\bm{x}^{\rm pt}_{n}}\right).
\end{equation}

\textbf{LGM in 3D Sparse Convolution.} 
According to the definition of 3D sparse convolution, it only conducts on existing voxels $\mathcal{X}$. If the voxel $\bm{x}_m$ does not contain point(s) and disappears after one attack step, the convolution on voxel $\bm{x}_m$ becomes invalid. This change in network architecture causes the learned perturbation may not be efficient to the changed new architecture. To be aware of the dynamic location of sparse convolution, we consider the changes of the occupancy values $o(\bm{x},\mathcal{X})$ along with the perturbations on the input. We present a dynamics-aware 3D sparse convolution and rewrite (\ref{conv3}) as:
\begin{equation}
\label{conv2}
\bm{f}_{\rm LGM}' = o(\bm{x},\mathcal{X})\cdot\left(\sum_{\bm{x}_q\in \mathcal{K}(\bm{x}, \mathcal{X}, \mathcal{S})}W_{\bm{x}_q-\bm{x}}\bm{f}_q + \bm{b}\right).
\end{equation}

Although $\bm{f}_{\rm FGM}'$ and $\bm{f}_{\rm LGM}'$ in (\ref{conv3}) and (\ref{conv2}) are the same in value, the formulas of gradient calculation are different. The partial derivative of $\bm{f}_{\rm LGM}'$ in (\ref{conv2}) with respect to a point $\bm{x}^{\rm pt}_{n}\in\mathcal{X}^{\rm pt}$ is:
\begin{equation}
\label{gradient2}
\begin{split}
\frac{\partial\bm{f}_{\rm LGM}'}{\partial\bm{x}^{\rm pt}_{n}} =& ~\bar{o}\cdot\left(\sum_{\bm{x}_q\in \mathcal{K}(\bm{x}, \mathcal{X}, \mathcal{S})}W_{\bm{x}_q-\bm{x}}\frac{\partial\bm{f}_q}{\partial\bm{x}^{\rm pt}_{n}}\right)\\ 
&+ \frac{\partial o(\bm{x},\mathcal{X})}{\partial\bm{x}^{\rm pt}_{n}} \cdot\left(\sum_{\bm{x}_q\in \mathcal{K}(\bm{x}, \mathcal{X}, \mathcal{S})}W_{\bm{x}_q-\bm{x}}\bm{f}_q + \bm{b}\right).
\end{split}
\end{equation}
Compared with traditional fast gradient calculation that does not consider dynamic architecture changes in (\ref{gradient}), we can discover that (\ref{gradient2}) has one more derivative on $\frac{\partial o(\bm{x},\mathcal{X})}{\partial\bm{x}^{\rm pt}_{n}}$, which ensures the existence change of convolution is reflected in the back-propagated gradients. 
Fig.~\ref{attack} illustrates the difference between sparse convolution, generalized sparse convolution, and our dynamics-aware sparse convolution. The figure is presented in the 2D view.


\textbf{Non-differentiable Function Approximation.}
To calculate the derivative $\frac{\partial o(\bm{x},\mathcal{X})}{\partial\bm{x}^{\rm pt}_{n}}$, it can be released by $\frac{\partial o(\bm{x},\mathcal{X})}{\partial\bm{x}}\frac{\partial \bm{x}}{\partial\bm{x}^{\rm pt}_{n}}$.
However, due to the binary definition of $o(\bm{x},\mathcal{X})\in\{0,1\}$ and the non-differentiable voxelization process in (\ref{vol}-\ref{vol2}), the derivative $\frac{\partial o(\bm{x},\mathcal{X})}{\partial\bm{x}}$ and $\frac{\partial \bm{x}}{\partial\bm{x}^{\rm pt}_{n}}$ are both not everywhere differentiable in the domain of definition, leading to the received gradient valueless. 
Following BPDA~\cite{athalye2018obfuscated}, we choose to reformulate the hard occupancy $o(\bm{x},\mathcal{X})$ into a differentiable soft occupancy $\hat{o}(\bm{x},\mathcal{X}^{\rm pt})\approx o(\bm{x},\mathcal{X})$ that directly requires point coordinates as input.
To model the soft occupancy function $\hat{o}(\bm{x},\mathcal{X}^{\rm pt})$, we consider the occupancy value of an arbitrary voxel $\bm{x}$ in 3D space is essentially determined by a conditional statement: “If at least one point in the point cloud belongs to the voxel, then the occupancy value is 1, and otherwise 0.” This process can be split into two stages: 
\begin{enumerate}[]
	\item Obtain a relation value $r(\bm{x}, \bm{x}^{\rm pt}_{n})\in\{0,1\}$ to show the existence of the point $\bm{x}^{\rm pt}_{n}$ in the given voxel $\bm{x}$.
	\item Gather all the relation values for the given voxel $\bm{x}$ together and find whether 1 exists. If at least one relation value equals 1, then the occupancy value of the given voxel is 1, and otherwise 0.
\end{enumerate}
Because both the two stages are not everywhere differentiable, we need to find functions to approximate them. 

\begin{figure}[t]
	\centering
	\subfigure[\scriptsize Original hard function]{
		\begin{minipage}{0.45\linewidth}
			\centering
			\includegraphics[width=1\textwidth, trim=0 -10 0 30,clip]{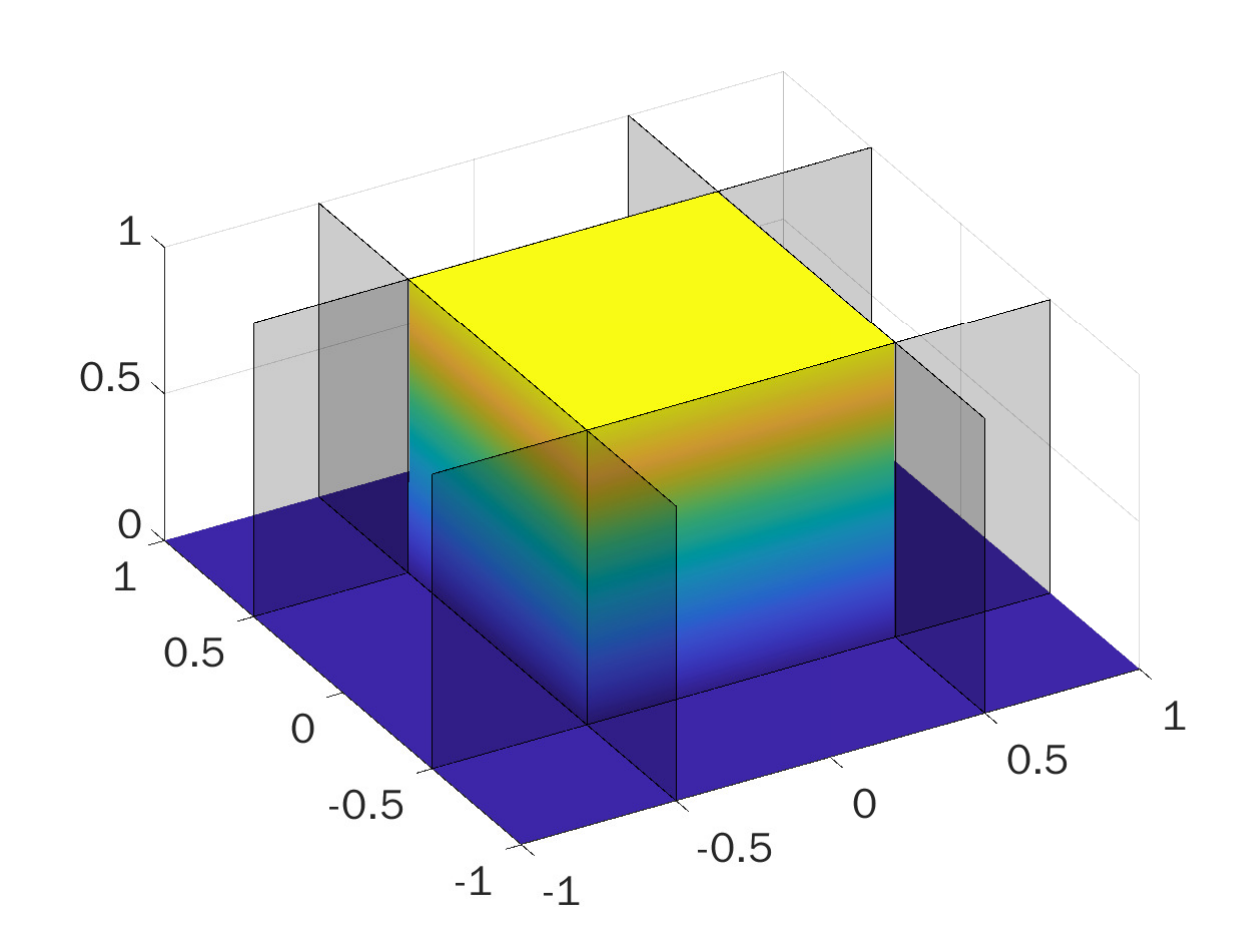}
		\end{minipage}
	}
	\subfigure[\scriptsize Bilinear interpolation~\cite{tu2020physically}]{
		\begin{minipage}{0.45\linewidth}
			\centering
			\includegraphics[width=1\textwidth, trim=0 -10 0 30,clip]{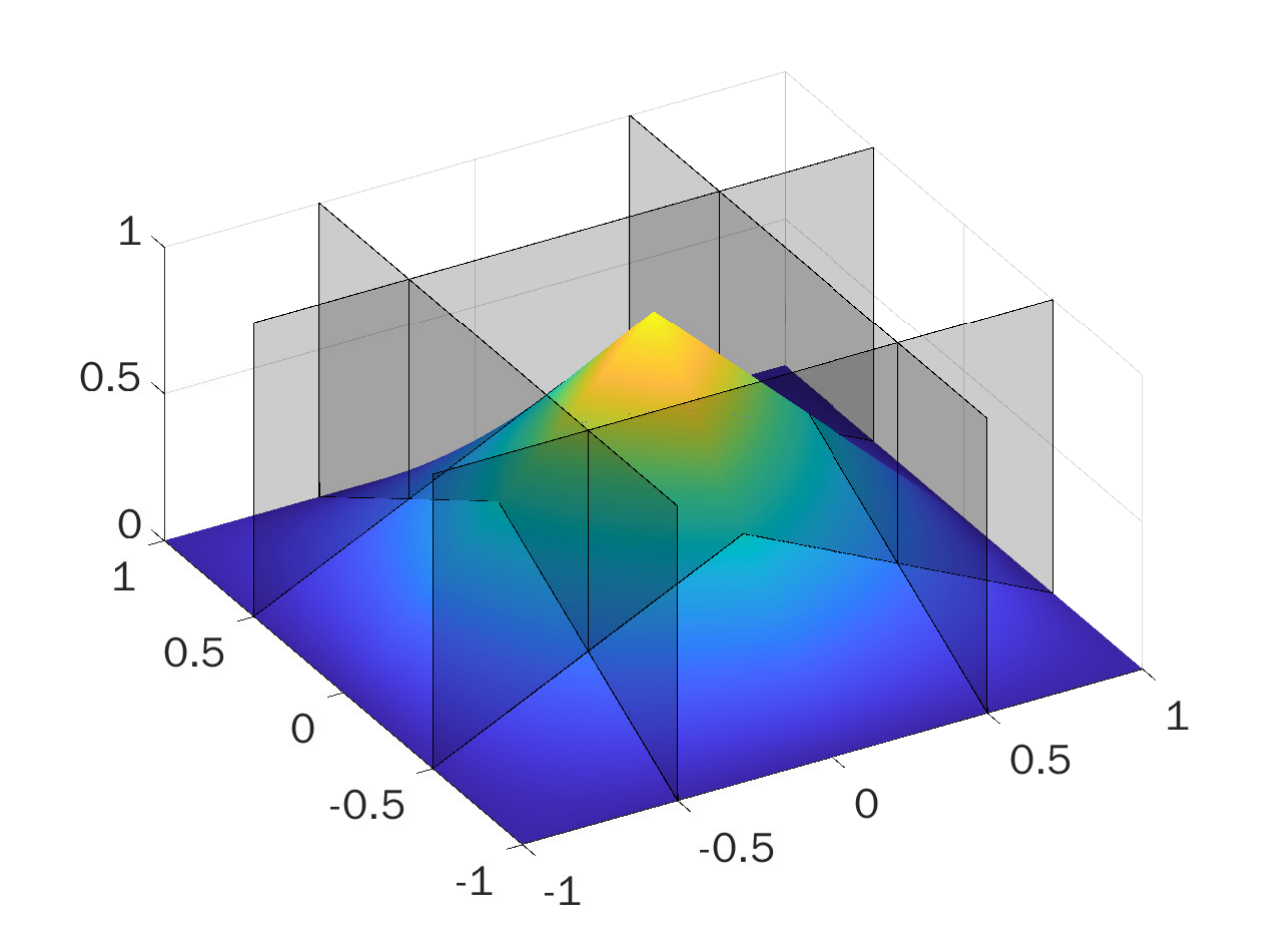}
		\end{minipage}
	}
	\subfigure[\scriptsize Radial basis function~\cite{qian2020end}]{
		\begin{minipage}{0.45\linewidth}
			\centering
			\includegraphics[width=1\textwidth, trim=0 -10 0 30,clip]{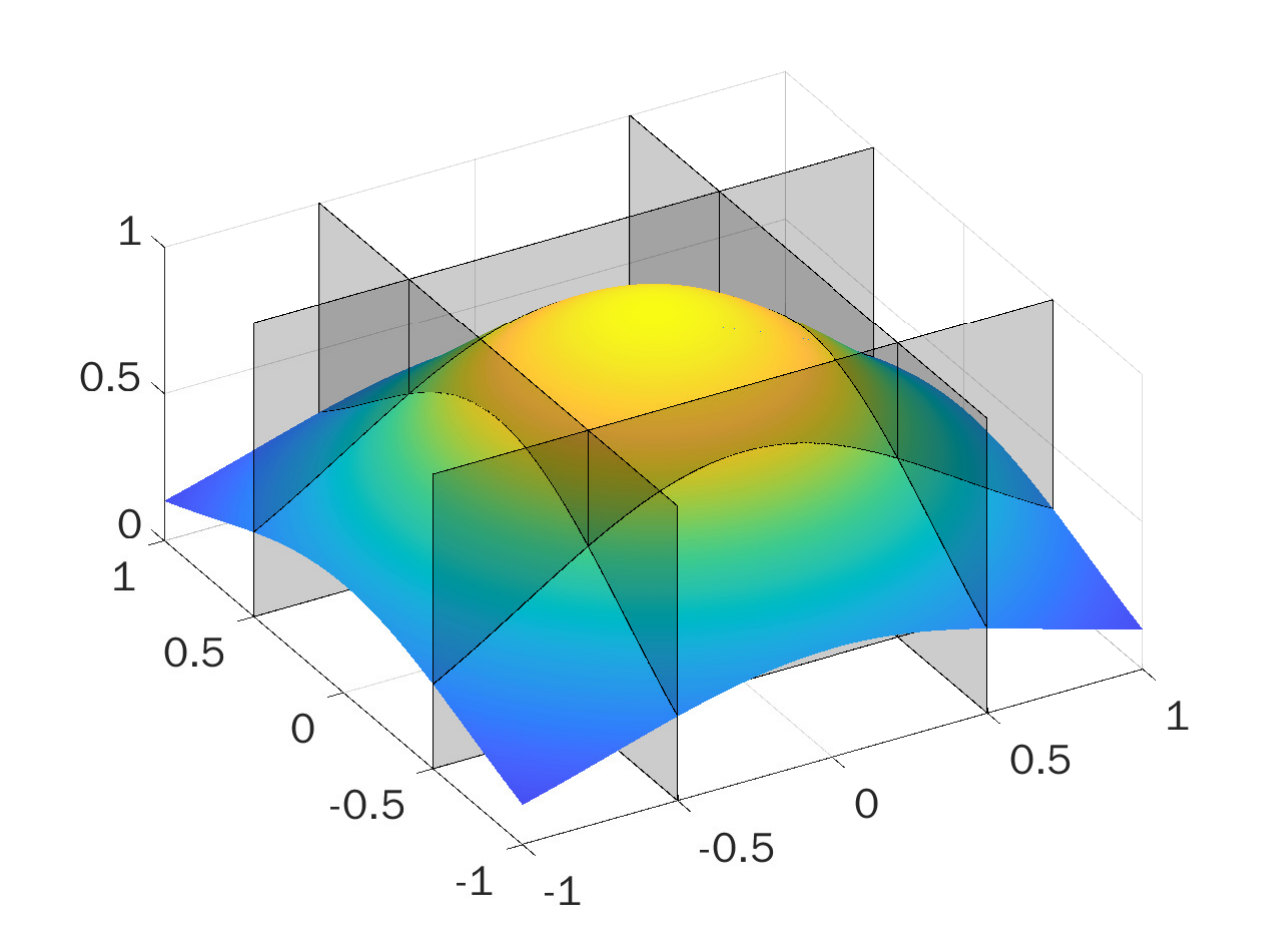}
		\end{minipage}
	}
	\subfigure[\scriptsize Sigmoid-like function]{
		\begin{minipage}{0.45\linewidth}
			\centering
			\includegraphics[width=1\textwidth, trim=0 -10 0 30,clip]{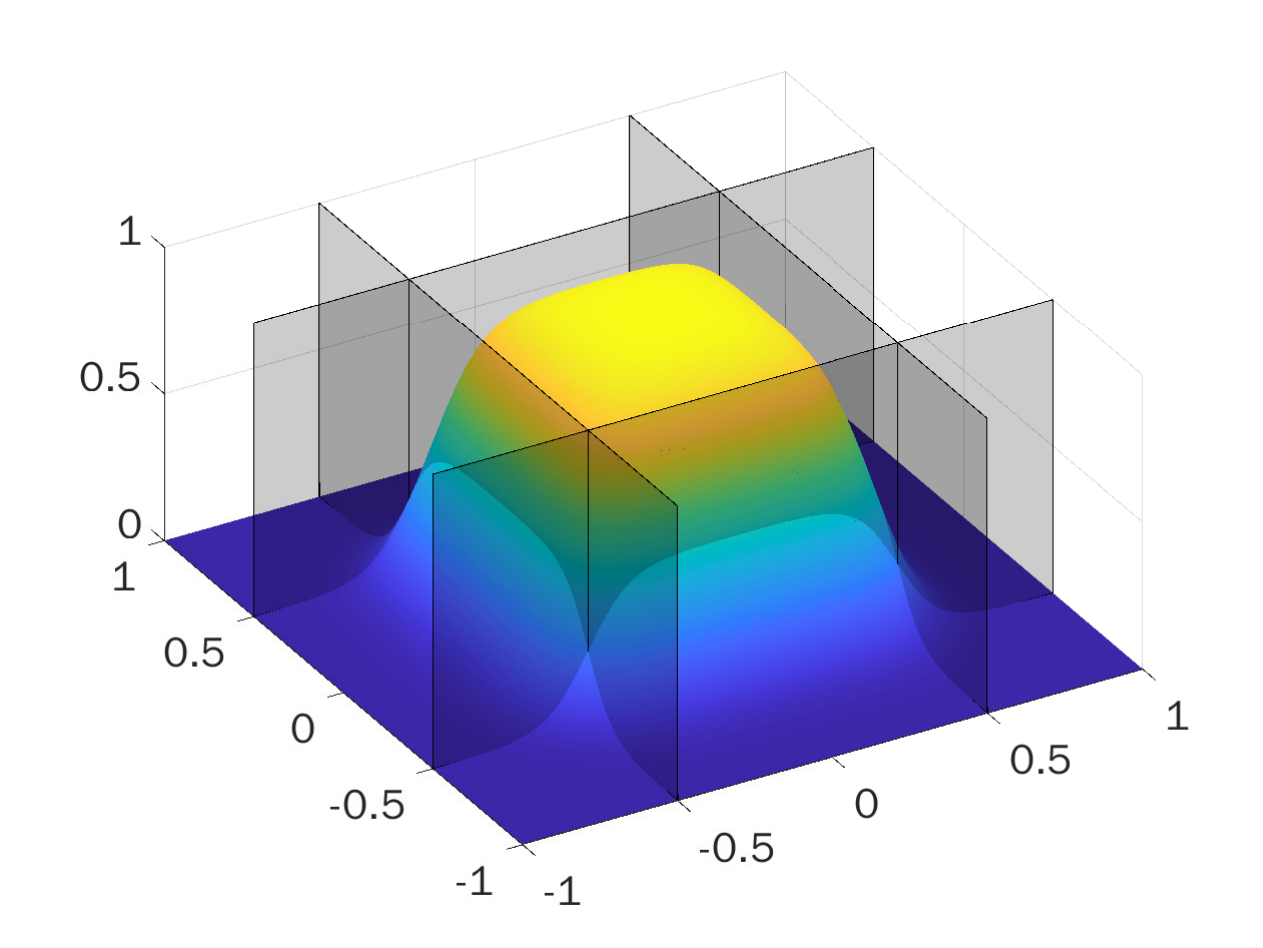}
		\end{minipage}
	}
	\caption{The visualization results on various relation functions. The gray planes show the voxel boundaries. Our sigmoid-like function well approximates the original function and has nice gradient quality near the voxel boundaries.}
	\label{sig_fig}
	\vspace{-10pt}
\end{figure}

For the first stage, we present a differentiable function $\hat{r}(\bm{x}, \bm{x}^{\rm pt}_{n})\in(0,1)$ to replace $r(\bm{x}, \bm{x}^{\rm pt}_{n})\in\{0,1\}$ for occupancy calculation. When the point $\bm{x}^{\rm pt}_{n}$ is nearer to the center of the voxel $\bm{x}$, we let the output relation value be higher and closer to 1. 
Some functions have been proposed to solve the non-differentiable problem in voxel occupancy, like bilinear interpolation~\cite{tu2020physically} and radial basis function~\cite{qian2020end}.
In this paper, we construct a sigmoid-like function and find it more suitable in attacks on 3D voxel-based networks. The differentiable relation function $\hat{r}(\bm{x}, \bm{x}^{\rm pt}_{n})$ is as follows:
\begin{gather}
\label{d}
d(\bm{x}, \bm{x}^{\rm pt}_{n}) = |(\bm{x}+0.5)-\bm{x}^{\rm pt}_{n}/L|,\\
\label{sig}
\hat{r}(\bm{x}, \bm{x}^{\rm pt}_{n}) = \prod_{i\in\{0,1,2\}}\frac{1}{1+{\rm exp}(\lambda\cdot(d(\bm{x}, \bm{x}^{\rm pt}_{n})_i-0.5))},
\end{gather}
where $d(\bm{x}, \bm{x}^{\rm pt}_{n})\in\mathbb{R}^3$ outputs a distance vector, $L$ is the voxel size, $d(\bm{x}, \bm{x}^{\rm pt}_{n})_i$ represents the $i$-th element of $d(\bm{x}, \bm{x}^{\rm pt}_{n})$, and $\lambda$ is a parameter that controls the slope near the voxel boundary. Fig.~\ref{sig_fig} shows the visualization results of original hard function, bilinear interpolation~\cite{tu2020physically}, radial basis function~\cite{qian2020end}, and our sigmoid-like function.
Compared to the existing two functions, our sigmoid-like function is more similar to the original hard function and has more significant gradient variation near voxel boundaries. 
When a point locates near the voxel boundary, the gradient will become distinctly large to force the point to quickly move in/out of the voxel, instead of staying near the boundary. 
We also conduct experiments on these functions and demonstrate the superior of our sigmoid-like function. 

For the second stage, we consider the gathering as an “or” operation in nature and present a differentiable function $g(\cdot)$ for this operation. Finally, our reformulated $\hat{o}(\bm{x},\mathcal{X}^{\rm pt})$ is derived as follows:
\begin{equation}
\label{o_hat}
\begin{split}
\hat{o}(\bm{x},\mathcal{X}^{\rm pt}) &= g(\{\hat{r}(\bm{x}, \bm{x}^{\rm pt}_{n})|\bm{x}^{\rm pt}_{n}\in \mathcal{N}(\bm{x}, \mathcal{X}^{\rm pt})\})\\
&= 1 - \prod_{\bm{x}^{\rm pt}_{n}\in \mathcal{N}(\bm{x}, \mathcal{X}^{\rm pt})}(1-\hat{r}(\bm{x}, \bm{x}^{\rm pt}_{n})),
\end{split}
\end{equation}
where $\mathcal{N}(\bm{x}, \mathcal{X}^{\rm pt})$ is a subset of point cloud $\mathcal{X}^{\rm pt}$ to reduce the huge computational costs. For a given voxel $\bm{x}$, we only gather the relation values of those points that possibly exist in the voxel after one attack step. If a voxel has no points to gather values, the voxel is ignored throughout the network. Therefore, we actually extend existing $M$ sparse voxels into $M'$ voxels, by adding a number of surrounding empty voxels that are possibly valid after one attack step. 
In our implementations, the step size at each iteration to perturb the point cloud coordinates is set very small for better optimizing performance. We find $M'\le 4M$. Theoretically $3\times3\times3-1=26$ convolutions are avaliable to activate around an existence sparse voxel (Generalized 3D Sparse Conv in Fig.~\ref{attack}), so the computational cost of our extended convolutions is tolerable. 
By multiplying the specially designed occupancy function on every convolution, we can obtain a leaded gradient that is aware of the dynamic network architecture.

\subsection{Summary}
Regardless of the forms of Adaptive Neural Networks, they are essentially constructed by input-adaptively computation units, which can be summarised as (\ref{f}). Given an input data $\bm{x}$, either 2D images, 3D point clouds, or other forms, each input-adaptive unit can be formulated as (\ref{o2}) with a binary occupancy function $o(\bm{x})\in\{0,1\}$. The problem of applying LGM is how to design the differentiable approximation function $\hat{o}(\bm{x})\in[0,1]$ to replace the non-differentiable function $o(\bm{x})$ in gradient back-propagation. 

In our work, we investigate the three most representative types of adaptive neural networks (layer skipping network~\cite{wang2018skipnet}, 2D sparse convolution network~\cite{verelst2020dynamic}, and 3D sparse convolution network~\cite{graham20183d}) which can cover almost all scenarios in vision adversarial attacks. In the subsection of Non-differentiable Function Approximation, we also propose three important criteria when designing $\hat{o}(\bm{x})$, which can provide necessary cues when designing dynamics-aware attacks on other rare forms of adaptive neural networks.

\section{Experiments on 2D Image Attack}
In this section, we evaluate our proposed dynamics-aware attack on layer skipping network and 2D sparse convolution network on 2D image datasets and analyze their performance against baseline dynamics-unaware methods. 

\subsection{Datasets and Implementation Details}

We conduct experiments on two popular image datasets: CIFAR-10~\cite{krizhevsky2009learning} and ImageNet~\cite{russakovsky2015imagenet}. 

%

We adopt SkipNet~\cite{wang2018skipnet} for layer skipping network and DynConv~\cite{verelst2020dynamic} for 2D sparse convolution network. In the experiments of CIFAR-10, the network backbone is ResNet-110~\cite{he2016deep} for SkipNet and ResNet-32~\cite{he2016deep} for DynConv. We also provide attack results on the normally trained static ResNet-110 and ResNet-32 for reference. On the ImageNet dataset, SkipNet and DynConv both adopt the network backbone ResNet-101~\cite{he2016deep} in their work, which enables us to fairly compare the attack performance between dynamically changing the network architecture layer-wise and within the layer. We also provide attack results on the normally trained static ResNet-101 for reference.

We perform our FGM and LGM on the basic attack algorithms FGSM~\cite{goodfellow2015explaining} to show the primitive attack performance difference between FGM and LGM in a single attack step and multiple attack steps. We constrain the maximum perturbation magnitude $\epsilon$ in $L_\infty$ norm. We also apply four representative iterative attack(BIM~\cite{kurakin2016adversarial}, PGD~\cite{madry2018towards}, C\&W~\cite{carlini2017towards}, and AutoAttack~\cite{croce2020reliable}) to further show the effectiveness of our LGM method against FGM. For all iterative attacks, we set the number of iteration as 100. For the hyper parameter $\lambda$ in LGM, we set it in range $0.0001\sim40$ for different attack settings. We find that given a large attack step $\alpha$, such as $\alpha=16$ in the single step attack FGSM, when the $\lambda$ is small our LGM performs the best. 

\begin{table*}[t]
	\caption{Adversarial attack results (\%) on the CIFAR-10 testing set in different $\epsilon$ for various models and methods.}
	\vspace{-5pt}
	{\footnotesize 
		\begin{center}
			\begin{tabular}{l|c|cc|cc|c|cc|cc|c}
				\toprule
				\multirow{3}{*}{Method} & \multirow{3}{*}{$\epsilon$} & \multicolumn{4}{c|}{SkipNet~\cite{wang2018skipnet} (ResNet-110~\cite{he2016deep})} & ResNet-110~\cite{he2016deep} & \multicolumn{4}{c|}{DynConv~\cite{verelst2020dynamic} (ResNet-32~\cite{he2016deep})} & ResNet-32~\cite{he2016deep} \\
				& & \multicolumn{2}{c|}{FGM} & \multicolumn{2}{c|}{LGM} & FGM & \multicolumn{2}{c|}{FGM} & \multicolumn{2}{c|}{LGM} & FGM \\
				& & ACC\textdownarrow & Arch. Chg. & ACC\textdownarrow & Arch. Chg. & ACC\textdownarrow & ACC\textdownarrow & Arch. Chg. & ACC\textdownarrow & Arch. Chg. & ACC\textdownarrow \\
				\midrule
				Bf. Attack & - & \multicolumn{4}{c|}{93.30} & 93.68 & \multicolumn{4}{c|}{93.05} & 92.63 \\
				\midrule
				FGSM~\cite{goodfellow2015explaining} & 1 & 26.70 & 4.64 & \textbf{25.52} & 4.47 & 36.00 & 42.11 & 1.75 & \textbf{39.54} & 1.76 & 31.33 \\
				FGSM~\cite{goodfellow2015explaining} & 2 & 16.67 & 8.04 & \textbf{12.23} & 8.17 & 25.77 & 30.63 & 2.71 & \textbf{18.13} & 2.70 & 21.45 \\
				FGSM~\cite{goodfellow2015explaining} & 4 & 11.76 & 12.16 & \textbf{6.52} & 12.68 & 20.61 & 22.92 & 3.87 & \textbf{6.57} & 3.87 & 16.76 \\
				FGSM~\cite{goodfellow2015explaining} & 8 &  9.69 & 17.18 & \textbf{6.71} & 17.26 & 16.03 & 14.08 & 5.36 & \textbf{5.05} & 5.27 & 12.01 \\
				\bottomrule
			\end{tabular}
		\end{center}
	}
	\label{cifar}
		\vspace{-5pt}
\end{table*}

\begin{table*}[t]
	\caption{Non-targeted and targeted adversarial attack results (\%) on the CIFAR-10 testing set for various models and methods.}
	\vspace{-5pt}
	{\footnotesize 
		\begin{center}
			\begin{tabular}{l|c|m{1cm}<{\centering}|m{0.4cm}<{\centering}m{1.3cm}<{\centering}|m{0.4cm}<{\centering}m{1.3cm}<{\centering}|m{0.4cm}<{\centering}m{1.3cm}<{\centering}|m{0.4cm}<{\centering}m{1.3cm}<{\centering}|c|c}
				\toprule
				\multirow{2}{*}{Method} & \multirow{2}{*}{$\epsilon$} & \multirow{2}{*}{Valid Pxl.} &  \multicolumn{2}{c|}{FGM} & \multicolumn{2}{c|}{FGM (Targeted)} & \multicolumn{2}{c|}{LGM} & \multicolumn{2}{c|}{LGM (Targeted)} & FGM & FGM (Targeted) \\
				& & & Suc.\textuparrow & Arch. Chg. & Suc.\textuparrow & Arch. Chg. & Suc.\textuparrow & Arch. Chg. & Suc.\textuparrow & Arch. Chg. & Suc.\textuparrow & Suc.\textuparrow \\
				\midrule
				& & & \multicolumn{8}{c|}{SkipNet~\cite{wang2018skipnet} (ResNet-110~\cite{he2016deep})} & \multicolumn{2}{c}{ResNet-110~\cite{he2016deep}} \\
				\midrule
				BIM~\cite{kurakin2016adversarial} & 8 & 100\% & 100.00 & 14.58 & 100.00 & 14.15 & 100.00 & 13.38 & 100.00 & 14.26 & 100.00 & 100.00 \\
				\midrule
				BIM~\cite{kurakin2016adversarial} & 8 & 5\% & ~52.62 & 3.04 & ~19.11 & 3.19 & ~\textbf{53.46} & 2.89 & ~\textbf{19.70} & 2.83 & 49.54 & 18.49 \\
				PGD~\cite{madry2018towards} & 8 & 5\% & ~51.20 & 3.10 & ~17.93 & 3.26 & ~\textbf{52.21} & 3.00 & ~\textbf{18.71} & 2.92 & 48.80 & 17.61  \\
				C\&W~\cite{carlini2017towards} & 8 & 5\% & ~53.60 & 3.08 & ~20.89 & 3.28 & ~\textbf{53.68} & 2.49 & ~\textbf{21.02} & 2.41 & 50.53 & 20.60 \\
				AutoAttack~\cite{croce2020reliable} & 8 & 5\% & ~54.02 & 3.05 & ~20.21 & 3.19 & ~\textbf{54.62} & 2.89 & ~\textbf{20.87} & 2.85 & 50.84 & 19.90 \\
				\midrule
				& & & \multicolumn{8}{c|}{DynConv~\cite{verelst2020dynamic} (ResNet-32~\cite{he2016deep})} & \multicolumn{2}{c}{ResNet-32~\cite{he2016deep}} \\
				\midrule
				BIM~\cite{kurakin2016adversarial} & 8 & 100\% & 100.00 & 4.94 & 100.00 & 4.71 & 100.00 & 5.13 & 100.00 & 4.71 & 100.00 & 100.00 \\
				\midrule
				BIM~\cite{kurakin2016adversarial} & 8 & 5\% & ~46.27 & 1.04 & ~15.79 & 0.99 & ~\textbf{46.93} & 1.04 & ~\textbf{16.24} & 0.98 & 56.83 & 22.01 \\
				PGD~\cite{madry2018towards} & 8 & 5\% & ~44.92 & 1.13 & ~15.07 & 1.10 & ~\textbf{45.95} & 1.13 & ~\textbf{15.48} & 1.09 & 55.39 & 20.85 \\
				C\&W~\cite{carlini2017towards} & 8 & 5\% & ~46.76 & 1.09 & ~16.78 & 1.03 & ~\textbf{47.00} & 1.05 & ~\textbf{17.09} & 0.99 & 57.94 & 24.14 \\
				AutoAttack~\cite{croce2020reliable} & 8 & 5\% & ~47.29 & 1.03 & ~16.63 & 0.98 & ~\textbf{47.85} & 1.03 & ~\textbf{16.93} & 0.98 & 58.30 & 23.37 \\
				\bottomrule
			\end{tabular}
		\end{center}
	}
	\label{cifar_t}
		\vspace{-10pt}
\end{table*}

\subsection{Results and Analysis}

Tables~\ref{cifar}, \ref{cifar_t} and \ref{imagenet} show the attack results on the CIFAR-10~\cite{krizhevsky2009learning} and  ImageNet~\cite{russakovsky2015imagenet} datasets in different maximum perturbation magnitudes $\epsilon$. “Bf. Attack” is “Before Attack” for short. “Valid Pxl.” is “Valid Pixel” for short, which denotes the percentage of attacked pixels in all pixels of the image. On all datasets with all $\epsilon$, our LGM significantly outperforms the dynamics-unaware FGM, which fully validates the importance of the attack method to be aware of the dynamic architecture changes. In the FGSM~\cite{goodfellow2015explaining} attack, the advantage of LGM is more apparent when $\epsilon\geq2$, which gives strong evidence of the superiority of our LGM. Since iterative attack can successfully attack all images when all pixels are perturbed, we restrict the percentage of pixels in attack. In Table~\ref{cifar_t}, we apply four representative iterative attack BIM~\cite{kurakin2016adversarial}, PGD~\cite{madry2018towards}, C\&W~\cite{carlini2017towards}, and AutoAttack~\cite{croce2020reliable} in both non-targeted and targeted scenario. In all experiments, we find our LGM superior to the dynamics-unaware baseline FGM, and our LGM attack can boost the performance of existing attacks on adaptive neural networks.

To evaluate the magnitude of network architecture change during the attack, we calculate the architecture change ratio as the number of changed computation units after the attack divided by the number of all changeable computation units. For the layer skipping network SkipNet~\cite{wang2018skipnet}, the architecture is changed layer-wise and the number of changeable layers is 53 for ResNet-110 and 32 for ResNet-101. For the 2D sparse convolution network DynConv~\cite{verelst2020dynamic}, the architecture is changed pixel-wise and the number of changeable pixels is the sum of the pixel number of the input image and all inner feature maps. In Tables~\ref{cifar} and \ref{imagenet}, we list the architecture change ratio (\%) as “Arch. Chg.”. We find that our LGM tends to change more computation units in FGSM~\cite{goodfellow2015explaining} attack. We initially thought that our LGM may tend to change more computation units to cause chaos in the architecture in order to misclassify the input image, but the results show that whether to change a computation unit should be appropriately based on its status.

Because the layer skipping network and 2D sparse convolution network deactivate unnecessary computation units in the static neural networks, we also compare their robustness against their corresponding static neural networks. We show the attack performance on ResNet-110 and ResNet-32 in Tables~\ref{cifar} and \ref{cifar_t} for the CIFAR-10 dataset, and ResNet-101 in Table~\ref{imagenet} for the ImageNet dataset. When adopting dynamics-unaware FGM in attack, the results reveal that the layer skipping network SkipNet~\cite{wang2018skipnet} is less robust than its corresponding static neural network, while the 2D sparse convolution network DynConv~\cite{verelst2020dynamic} shows better robustness. When adopting LGM in the FGSM~\cite{goodfellow2015explaining} attack with $\epsilon\geq2$, the adaptive neural networks are all easier to be attacked than static neural networks.

\begin{table*}[t]
	\caption{Adversarial attack results (\%) on the ImageNet validation set in different $\epsilon$ for various models and methods.}
	\vspace{-3pt}
	{\footnotesize 
		\begin{center}
			\begin{tabular}{l|c|cc|cc|cc|cc|c}
				\toprule
				\multirow{3}{*}{Method} & \multirow{3}{*}{$\epsilon$} & \multicolumn{4}{c|}{SkipNet~\cite{wang2018skipnet} (ResNet-101~\cite{he2016deep})} & \multicolumn{4}{c|}{DynConv~\cite{verelst2020dynamic} (ResNet-101~\cite{he2016deep})} & ResNet-101~\cite{he2016deep}\\
				& & \multicolumn{2}{c|}{FGM} & \multicolumn{2}{c|}{LGM} & \multicolumn{2}{c|}{FGM} & \multicolumn{2}{c|}{LGM} & FGM \\
				& & ~mIoU\textdownarrow~ & Arch. Chg. & ~mIoU\textdownarrow~ & Arch. Chg. & ~mIoU\textdownarrow~ & Arch. Chg. & ~mIoU\textdownarrow~ & Arch. Chg. & ~mIoU\textdownarrow~ \\
				\midrule
				Bf. Attack~~~~ & - & \multicolumn{4}{c|}{77.02} & \multicolumn{4}{c|}{76.85} & 77.31 \\
				\midrule
				FGSM~\cite{goodfellow2015explaining} & 1 & 10.14 & 3.15 & \textbf{10.09} & 3.42 & 11.69 & 2.65 & \textbf{11.31} & 2.77 & 9.29 \\
				FGSM~\cite{goodfellow2015explaining} & 2 & 9.45 & 4.39 & \textbf{6.82} & 4.63 & 10.20 & 3.89 & \textbf{7.84} & 4.13 & 8.39 \\
				FGSM~\cite{goodfellow2015explaining} & 4 & 11.61 & 5.87 & \textbf{5.35} & 6.08 & 11.60 & 5.53 & \textbf{4.52} & 5.82 & 10.14 \\
				FGSM~\cite{goodfellow2015explaining} & 8 & 15.33 & 7.86 & \textbf{5.25} & 8.10 & 14.37 & 7.37 & \textbf{3.36} & 7.66 & 12.89 \\
				FGSM~\cite{goodfellow2015explaining} & 16 & 12.28 & 10.81 & \textbf{4.28} & 10.92 & 11.31 & 9.71 & \textbf{2.54} & 10.02 & 9.67 \\
				\bottomrule
			\end{tabular}
		\end{center}
	}
	\label{imagenet}
\end{table*}

\begin{figure*}[t]
	\vspace{-10pt}
	\begin{center}
		\includegraphics[width=1\linewidth, trim=0 0 0 0,clip]{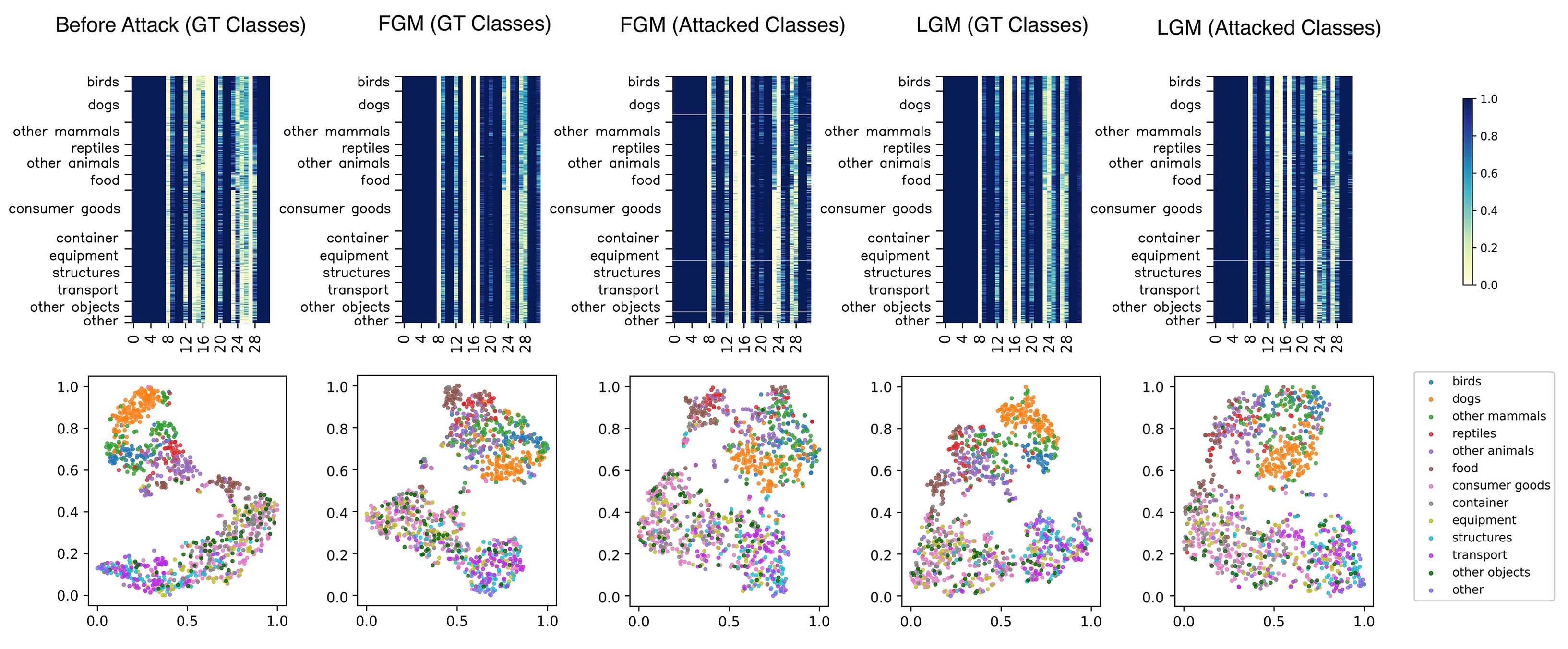}
	\end{center}
	\vspace{-5pt}
	~~~~~~~~~\,~~~~~~~{\footnotesize (a)}~~~~~~~~~~~~~~~~~~~~~~~{\footnotesize (b)}~~~~~~~~~~~~~~~~~~~~~~~{\footnotesize (c)}~~~~~~~~~~~~~~~~~~~~~~~{\footnotesize (d)}~~~~~~~~~~~~~~~~~~~~~~~{\footnotesize (e)}
	\caption{
		The visualization results of layer-wise behavior in the layer skipping network before and after the attack on the ImageNet validation set. The top row is the rates at which different layers are executed (x-axis) for images of the 1000 classes in ImageNet (y-axis). The bottom row is the T-SNE results of the layer execution rates in top row. We conduct the BIM~\cite{kurakin2016adversarial} attack with maximum perturbation $\epsilon=16$. We attack for 100 iterations and all image pixels are valid.
	}
	\label{imagenet_img}
	\vspace{-5pt}
\end{figure*}

We also analyze the layer-wise activation behavior in the layer skipping network SkipNet~\cite{wang2018skipnet} before and after the attack. In Fig.~\ref{imagenet_img} (top), we follow \cite{almahairi2016dynamic} to show the rates at which different layers are executed (x-axis) for images of the 1000 classes in ImageNet (y-axis). The number of changeable layers is 32. “GT Classes” denotes the images presented as their ground-truth classes, and “Attacked Classes” denotes the images presented as their misclassified classes. We also show the mid-level categories on the y-axis for clarity. In Fig.~\ref{imagenet_img} (bottom), we present the T-SNE results of the layer execution rates in top row. Combining the histogram and T-SNE results before attack in Fig.~\ref{imagenet_img} (a), we observe a clear difference between layers used for man-made objects (\textit{food}, \textit{consumer goods}, \textit{container}, \textit{equipment}, \textit{structures}, \textit{transport}, and \textit{other objects}) and for animals (\textit{birds}, \textit{dogs}, \textit{other mammals}, \textit{reptiles}, and \textit{other animals}). We also find some mid-level categories have distinct differences in both the histogram and T-SNE results, such as \textit{birds}, \textit{dogs}, \textit{food}, and \textit{transport}. When the attack is finished in Figs.~\ref{imagenet_img} (b) and (d), we find that the layer-wise behaviors tend to be the same. In the histograms of Figs.~\ref{imagenet_img} (c) and (e), the white lines denote the images in these classes are all successfully attacked. 
The layer-wise behavior of mid-classes reveals that the layers can be selectively activated based on the semantic class of the input image. The semantic behavior of layers can also be kept to a certain degree after the attack. 
Compared to FGM, LGM can force more classes to be successfully attacked and tends to disrupt the layer-wise semantic behavior more. 

\section{Experiments on 3D Point Cloud Attack}
In this section, we evaluate our proposed dynamics-aware attack on 3D sparse convolution network on 3D point cloud datasets and analyze its performance against baseline dynamics-unaware methods. 

\begin{figure*}[t]
	\begin{center}
		\includegraphics[width=1\linewidth, trim=0 0 10 0,clip]{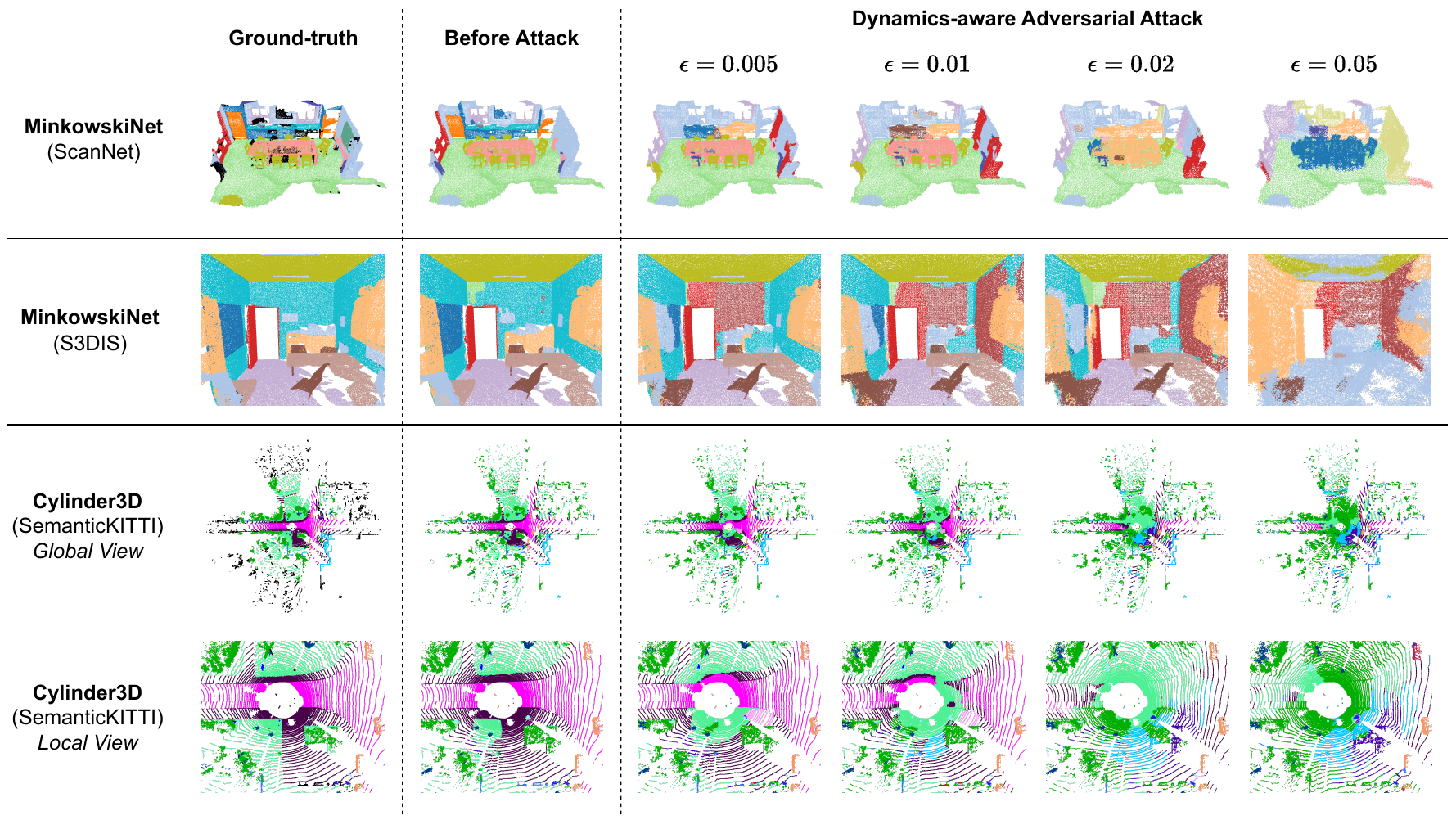}
	\end{center}
	\vspace{-12pt}
	\caption{The qualitative visualization results of LGM on the ScanNet, S3DIS and SemanticKITTI dataset in different $\epsilon$ (m). The black areas in ground-truth are unlabeled.}
	\vspace{-7pt}
	\label{visual}
\end{figure*}

\begin{table*}[t]
	\footnotesize
	\caption{Adversarial attack results (\%) on 3D point cloud datasets in different $\epsilon$ for various methods.}
	\vspace{-3pt}
	{\footnotesize 
		\begin{center}
			\begin{tabular}{l|c|cc|cc|cc}
				\toprule
				\multirow{2}{*}{Method} & ~~~~~\multirow{2}{*}{$\epsilon$}~~~~~ & \multicolumn{2}{c|}{MinkowskiNet~\cite{choy20194d} \textit{(ScanNet)}} & \multicolumn{2}{c|}{MinkowskiNet~\cite{choy20194d} \textit{(S3DIS)}} & \multicolumn{2}{c}{Cylinder3D~\cite{zhu2021cylindrical} \textit{(SemanticKITTI)}} \\
				& & ~~FGM mIoU\textdownarrow~ & ~LGM mIoU\textdownarrow~~ & ~FGM mIoU\textdownarrow~ & ~LGM mIoU\textdownarrow~ & ~~FGM mIoU\textdownarrow~ & ~LGM mIoU\textdownarrow~~ \\
				\midrule
				Bf. Attack & - & \multicolumn{2}{c|}{72.22} & \multicolumn{2}{c|}{65.47} & \multicolumn{2}{c}{66.91}\\
				\midrule
				Point-BIM~\cite{kurakin2016adversarial}~~~~ & 0.005 & 60.44 & \textbf{24.35} & 57.53 & \textbf{48.20} & 34.43 & \textbf{33.63} \\
				Point-BIM~\cite{kurakin2016adversarial} & 0.01 & 55.51 & \textbf{10.87} & 52.35 & \textbf{39.65} & 29.77 & \textbf{29.13} \\
				Point-BIM~\cite{kurakin2016adversarial} & 0.02 & 38.65 & \textbf{5.09} & 45.24 & \textbf{30.93} & 16.01 & \textbf{15.17} \\
				Point-BIM~\cite{kurakin2016adversarial} & 0.05 & 8.70 & \textbf{2.78} & 21.21 & \textbf{7.45} & 9.17 & \textbf{8.50} \\
				\bottomrule
			\end{tabular}
		\end{center}
	}
	\label{3d}
	\vspace{-10pt}
\end{table*}

\subsection{Datasets and Implementation Details}

We conduct experiments on three popular point cloud datasets, including two indoor scene datasets ScanNet~\cite{dai2017scannet}, S3DIS~\cite{armeni20163d,armeni2017joint}, and an outdoor scene dataset SemanticKITTI~\cite{behley2019semantickitti}. 

We adopt MinkowskiNet~\cite{choy20194d} on the ScanNet and S3DIS datasets, and Cylinder3D~\cite{zhu2021cylindrical} on the SemanticKITTI dataset for semantic segmentation. The two network architectures are both built on the base of 3D sparse convolution~\cite{graham20183d} and achieve state-of-the-art performance.

To attack the network, we compute a cross-entropy loss for gradient back-propagation. We adopt our FGM and LGM based on a variant version of BIM~\cite{kurakin2016adversarial} proposed by Liu \textit{et al.}~\cite{liu2019extending} that tailors the real number nature of point cloud coordinates. We denote it as Point-BIM in this paper. We constrain the maximum perturbation magnitude $\epsilon$ for each point onto the surface of a 3D cube by $L_\infty$ norm. The unit of $\epsilon$ is meter. We traverse different attack step size $\alpha$ in range $0.0005\sim0.025$ and different $\lambda$ for the sigmoid-like function (\ref{sig}) in range $10\sim55$ to adopt their best one for different $\epsilon$. 
The iteration number for the ScanNet and S3DIS datasets is 50, and for the SemanticKITTI dataset is 30.



\subsection{Results and Analysis}



Table~\ref{3d} show the attack results on the ScanNet~\cite{dai2017scannet}, S3DIS~\cite{armeni20163d,armeni2017joint}, and SemanticKITTI~\cite{behley2019semantickitti} datasets in different maximum perturbation magnitudes $\epsilon$. “Bf. Attack” is “Before Attack” for short. On all datasets with all $\epsilon$, our LGM significantly outperforms the dynamics-unaware FGM, which fully validates the importance of the attack method to be aware of the dynamic architecture changes. 
We find that the performance drop for LGM on the SemanticKITTI dataset is not as good as the ScanNet and S3DIS datasets. This phenomenon can be explained by the fact that the point cloud coordinates of the SemanticKITTI dataset are included in point features, while the point features of the ScanNet and S3DIS datasets only contain RGB colors. 
With the included point cloud coordinates, the gradient back-propagation pass has better quality than those point cloud features that do not include coordinates. 


Fig.~\ref{visual} show the visualization results of our dynamics-aware LGM on the ScanNet, S3DIS, and SemanticKITTI datasets. We observe that the perturbed point cloud coordinates are imperceptible to human eyes in some small $\epsilon$, such as $\epsilon=0.005$ m (0.5 cm) and $\epsilon=0.01$ m (1 cm) on the ScanNet and S3DIS datasets and $\epsilon=0.01$ m (1 cm) and $\epsilon=0.02$ m (2 cm) on the SemanticKITTI dataset, but these maximum perturbations do cause significant wrong predictions of the victim networks. This huge contrast reveals a great security risk in real-world applications of 3D sparse convolution networks.

We also conduct attacks that utilize different soft occupancy functions $\hat{o}(\bm{x}, \mathcal{X}^{\rm pt})$, including bilinear interpolation (BI)~\cite{tu2020physically}, radial basis function (RBF)~\cite{qian2020end}, and our sigmoid-like function in (\ref{sig}). Table~\ref{sig_ablation} shows the attack performance on the ScanNet validation set in different $\epsilon$. The results validate the effectiveness of our sigmoid-like function (denoted as Sigmoid). 
Compared to the visualized functions in Fig.~\ref{sig_fig}, we find that our sigmoid-like function imitates the hard occupancy better than RBF and BI, and their attack results in Table~\ref{sig_ablation} are consistent with their imitation qualities.

\begin{table}[t]
	\footnotesize
	\caption{Point cloud semantic segmentation mIoU results (\%) on the ScanNet validation set in different $\epsilon$ (m) for various occupancy functions.}
	\vspace{-5pt}
	{\footnotesize 
		\begin{center}
			\begin{tabular}{l|cccc}
				\toprule
				Method & ~$\epsilon=0.005$~ & ~$\epsilon=0.01$~ & ~$\epsilon=0.02$~ & ~$\epsilon=0.05$~ \\
				\midrule
				Bf. Attack~~ &\multicolumn{4}{c}{72.22}\\
				\midrule
				BI~\cite{tu2020physically} & 68.08 & 60.21 & 32.62 & 11.73\\
				RBF~\cite{qian2020end} & 36.14 & 16.62 & 8.29 & 5.00\\
				Sigmoid & \textbf{25.79} & \textbf{11.51} & \textbf{5.76} & \textbf{3.83}\\
				\bottomrule
			\end{tabular}
		\end{center}
	}
	\label{sig_ablation}
	\vspace{-10pt}
\end{table}

\section{Conclusion}
In this paper, we have investigated the lagged gradient issue in adversarial attacks for adaptive neural networks that have dynamic architecture and we have proposed a Leaded Gradient Method (LGM) for the dynamics-aware adversarial attack. We take layer skipping network and 2D/3D sparse convolution network as typical examples of adaptive neural networks to design our method. Specifically, we first analyze the missing part in traditional gradient calculation that considers the architecture changes. We then reformulate the gradient to better “lead” each attack step. Experimental results on both 2D images and 3D point clouds show our LGM achieves impressive performance and outperforms dynamics-unaware baseline methods. We believe our dynamics-aware attack can be used for other adaptive neural networks.

\section{Acknowledgement}
We thank He Wang, Ziyi Wu, Pengliang Ji, and Haowen Sun for their valuable supports on algorithms and experiments of adversarial attack against 3D sparse convolution networks.

\ifCLASSOPTIONcaptionsoff
\newpage
\fi



%
%
%

\bibliographystyle{IEEEtran}
\bibliography{egbib.bib}

\begin{thebibliography}{10}
\providecommand{\url}[1]{#1}
\csname url@samestyle\endcsname
\providecommand{\newblock}{\relax}
\providecommand{\bibinfo}[2]{#2}
\providecommand{\BIBentrySTDinterwordspacing}{\spaceskip=0pt\relax}
\providecommand{\BIBentryALTinterwordstretchfactor}{4}
\providecommand{\BIBentryALTinterwordspacing}{\spaceskip=\fontdimen2\font plus
\BIBentryALTinterwordstretchfactor\fontdimen3\font minus
  \fontdimen4\font\relax}
\providecommand{\BIBforeignlanguage}[2]{{%
\expandafter\ifx\csname l@#1\endcsname\relax
\typeout{** WARNING: IEEEtran.bst: No hyphenation pattern has been}%
\typeout{** loaded for the language `#1'. Using the pattern for}%
\typeout{** the default language instead.}%
\else
\language=\csname l@#1\endcsname
\fi
#2}}
\providecommand{\BIBdecl}{\relax}
\BIBdecl

\bibitem{szegedy2014intriguing}
C.~Szegedy, W.~Zaremba, I.~Sutskever, J.~Bruna, D.~Erhan, I.~Goodfellow, and
  R.~Fergus, ``Intriguing properties of neural networks,'' in \emph{Int. Conf.
  Learn. Represent.}, 2014, pp. 1--10.

\bibitem{goodfellow2015explaining}
I.~J. Goodfellow, J.~Shlens, and C.~Szegedy, ``Explaining and harnessing
  adversarial examples,'' in \emph{Int. Conf. Learn. Represent.}, 2015, pp.
  1--11.

\bibitem{moosavi2016deepfool}
S.-M. Moosavi-Dezfooli, A.~Fawzi, and P.~Frossard, ``Deepfool: a simple and
  accurate method to fool deep neural networks,'' in \emph{IEEE Conf. Comput.
  Vis. Pattern Recog.}, 2016, pp. 2574--2582.

\bibitem{carlini2017towards}
N.~Carlini and D.~Wagner, ``Towards evaluating the robustness of neural
  networks,'' in \emph{IEEE Symposium on Security and Privacy}, 2017, pp.
  39--57.

\bibitem{wang2018skipnet}
X.~Wang, F.~Yu, Z.-Y. Dou, T.~Darrell, and J.~E. Gonzalez, ``{SkipNet}:
  Learning dynamic routing in convolutional networks,'' in \emph{Eur. Conf.
  Comput. Vis.}, 2018, pp. 409--424.

\bibitem{verelst2020dynamic}
T.~Verelst and T.~Tuytelaars, ``Dynamic convolutions: Exploiting spatial
  sparsity for faster inference,'' in \emph{IEEE Conf. Comput. Vis. Pattern
  Recog.}, 2020, pp. 2320--2329.

\bibitem{graham20183d}
B.~Graham, M.~Engelcke, and L.~Van Der~Maaten, ``{3D} semantic segmentation
  with submanifold sparse convolutional networks,'' in \emph{IEEE Conf. Comput.
  Vis. Pattern Recog.}, 2018, pp. 9224--9232.

\bibitem{krizhevsky2009learning}
A.~Krizhevsky \emph{et~al.}, ``Learning multiple layers of features from tiny
  images,'' \emph{Tech. rep.}, 2009.

\bibitem{russakovsky2015imagenet}
O.~Russakovsky, J.~Deng, H.~Su, J.~Krause, S.~Satheesh, S.~Ma, Z.~Huang,
  A.~Karpathy, A.~Khosla, M.~Bernstein \emph{et~al.}, ``Imagenet large scale
  visual recognition challenge,'' \emph{Int. J. Comput. Vis.}, vol. 115, no.~3,
  pp. 211--252, 2015.

\bibitem{dai2017scannet}
A.~Dai, A.~X. Chang, M.~Savva, M.~Halber, T.~Funkhouser, and M.~Nie{\ss}ner,
  ``{ScanNet}: Richly-annotated {3D} reconstructions of indoor scenes,'' in
  \emph{IEEE Conf. Comput. Vis. Pattern Recog.}, 2017, pp. 5828--5839.

\bibitem{armeni20163d}
I.~Armeni, O.~Sener, A.~R. Zamir, H.~Jiang, I.~Brilakis, M.~Fischer, and
  S.~Savarese, ``{3D} semantic parsing of large-scale indoor spaces,'' in
  \emph{IEEE Conf. Comput. Vis. Pattern Recog.}, 2016, pp. 1534--1543.

\bibitem{armeni2017joint}
I.~Armeni, S.~Sax, A.~R. Zamir, and S.~Savarese, ``Joint {2D-3D}-semantic data
  for indoor scene understanding,'' \emph{arXiv preprint arXiv:1702.01105},
  2017.

\bibitem{behley2019semantickitti}
J.~Behley, M.~Garbade, A.~Milioto, J.~Quenzel, S.~Behnke, C.~Stachniss, and
  J.~Gall, ``Semantickitti: A dataset for semantic scene understanding of lidar
  sequences,'' in \emph{Int. Conf. Comput. Vis.}, 2019, pp. 9297--9307.

\bibitem{kurakin2016adversarial}
A.~Kurakin, I.~Goodfellow, S.~Bengio \emph{et~al.}, ``Adversarial examples in
  the physical world,'' in \emph{Int. Conf. Learn. Represent.}, 2017, pp.
  1--10.

\bibitem{madry2018towards}
A.~Madry, A.~Makelov, L.~Schmidt, D.~Tsipras, and A.~Vladu, ``Towards deep
  learning models resistant to adversarial attacks,'' in \emph{Int. Conf.
  Learn. Represent.}, 2018, pp. 1--10.

\bibitem{papernot2016limitations}
N.~Papernot, P.~McDaniel, S.~Jha, M.~Fredrikson, Z.~B. Celik, and A.~Swami,
  ``The limitations of deep learning in adversarial settings,'' in \emph{IEEE
  European Symposium on Security and Privacy}, 2016, pp. 372--387.

\bibitem{li2021simple}
Z.~Li, Y.~Shi, J.~Gao, S.~Wang, B.~Li, P.~Liang, and W.~Hu, ``A simple and
  strong baseline for universal targeted attacks on siamese visual tracking,''
  \emph{IEEE Transactions on Circuits and Systems for Video Technology},
  vol.~32, no.~6, pp. 3880--3894, 2021.

\bibitem{chen2017zoo}
P.-Y. Chen, H.~Zhang, Y.~Sharma, J.~Yi, and C.-J. Hsieh, ``{ZOO}: Zeroth order
  optimization based black-box attacks to deep neural networks without training
  substitute models,'' in \emph{ACM Workshop on Artificial Intelligence and
  Security}, 2017, pp. 15--26.

\bibitem{su2019one}
J.~Su, D.~V. Vargas, and K.~Sakurai, ``One pixel attack for fooling deep neural
  networks,'' \emph{IEEE Transactions on Evolutionary Computation}, vol.~23,
  no.~5, pp. 828--841, 2019.

\bibitem{dong2018boosting}
Y.~Dong, F.~Liao, T.~Pang, H.~Su, J.~Zhu, X.~Hu, and J.~Li, ``Boosting
  adversarial attacks with momentum,'' in \emph{IEEE Conf. Comput. Vis. Pattern
  Recog.}, 2018, pp. 9185--9193.

\bibitem{xiang2019generating}
C.~Xiang, C.~R. Qi, and B.~Li, ``Generating {3D} adversarial point clouds,'' in
  \emph{IEEE Conf. Comput. Vis. Pattern Recog.}, 2019, pp. 9136--9144.

\bibitem{liu2019extending}
D.~Liu, R.~Yu, and H.~Su, ``Extending adversarial attacks and defenses to deep
  {3D} point cloud classifiers,'' in \emph{IEEE Int. Conf. Image Process.},
  2019, pp. 2279--2283.

\bibitem{zheng2019pointcloud}
T.~Zheng, C.~Chen, J.~Yuan, B.~Li, and K.~Ren, ``Pointcloud saliency maps,'' in
  \emph{Int. Conf. Comput. Vis.}, 2019, pp. 1598--1606.

\bibitem{liu2020adversarial}
D.~Liu, R.~Yu, and H.~Su, ``Adversarial shape perturbations on {3D} point
  clouds,'' in \emph{Eur. Conf. Comput. Vis.}, 2020, pp. 88--104.

\bibitem{wen2020geometry}
Y.~Wen, J.~Lin, K.~Chen, C.~P. Chen, and K.~Jia, ``Geometry-aware generation of
  adversarial point clouds,'' \emph{IEEE Trans. Pattern Anal. Mach. Intell.},
  2020.

\bibitem{haque2020ilfo}
M.~Haque, A.~Chauhan, C.~Liu, and W.~Yang, ``{ILFO}: Adversarial attack on
  adaptive neural networks,'' in \emph{IEEE Conf. Comput. Vis. Pattern Recog.},
  2020, pp. 14\,264--14\,273.

\bibitem{hong2021panda}
S.~Hong, Y.~Kaya, I.-V. Modoranu, and T.~Dumitras, ``A panda? no, it's a sloth:
  Slowdown attacks on adaptive multi-exit neural network inference,'' in
  \emph{Int. Conf. Learn. Represent.}, 2021, pp. 1--17.

\bibitem{chen2022nicgslowdown}
S.~Chen, Z.~Song, M.~Haque, C.~Liu, and W.~Yang, ``{NICGSlowDown}: Evaluating
  the efficiency robustness of neural image caption generation models,'' in
  \emph{IEEE Conf. Comput. Vis. Pattern Recog.}, 2022, pp. 15\,365--15\,374.

\bibitem{krizhevsky2012imagenet}
A.~Krizhevsky, I.~Sutskever, and G.~E. Hinton, ``Imagenet classification with
  deep convolutional neural networks,'' \emph{Adv. Neural Inform. Process.
  Syst.}, vol.~25, 2012.

\bibitem{simonyan2015very}
K.~Simonyan and A.~Zisserman, ``Very deep convolutional networks for
  large-scale image recognition,'' in \emph{Int. Conf. Learn. Represent.},
  2015, pp. 1--14.

\bibitem{szegedy2015going}
C.~Szegedy, W.~Liu, Y.~Jia, P.~Sermanet, S.~Reed, D.~Anguelov, D.~Erhan,
  V.~Vanhoucke, and A.~Rabinovich, ``Going deeper with convolutions,'' in
  \emph{IEEE Conf. Comput. Vis. Pattern Recog.}, 2015, pp. 1--9.

\bibitem{he2016deep}
K.~He, X.~Zhang, S.~Ren, and J.~Sun, ``Deep residual learning for image
  recognition,'' in \emph{IEEE Conf. Comput. Vis. Pattern Recog.}, 2016, pp.
  770--778.

\bibitem{vaswani2017attention}
A.~Vaswani, N.~Shazeer, N.~Parmar, J.~Uszkoreit, L.~Jones, A.~N. Gomez,
  {\L}.~Kaiser, and I.~Polosukhin, ``Attention is all you need,'' \emph{Adv.
  Neural Inform. Process. Syst.}, vol.~30, 2017.

\bibitem{han2021dynamic}
Y.~Han, G.~Huang, S.~Song, L.~Yang, H.~Wang, and Y.~Wang, ``Dynamic neural
  networks: A survey,'' \emph{IEEE Trans. Pattern Anal. Mach. Intell.},
  vol.~44, no.~11, pp. 7436--7456, 2021.

\bibitem{bolukbasi2017adaptive}
T.~Bolukbasi, J.~Wang, O.~Dekel, and V.~Saligrama, ``Adaptive neural networks
  for efficient inference,'' in \emph{Int. Conf. Mach. Learn.}, 2017, pp.
  527--536.

\bibitem{veit2018convolutional}
A.~Veit and S.~Belongie, ``Convolutional networks with adaptive inference
  graphs,'' in \emph{Eur. Conf. Comput. Vis.}, 2018, pp. 3--18.

\bibitem{bengio2013estimating}
Y.~Bengio, N.~L{\'e}onard, and A.~Courville, ``Estimating or propagating
  gradients through stochastic neurons for conditional computation,''
  \emph{arXiv preprint arXiv:1308.3432}, 2013.

\bibitem{herrmann2020channel}
C.~Herrmann, R.~S. Bowen, and R.~Zabih, ``Channel selection using gumbel
  softmax,'' in \emph{Eur. Conf. Comput. Vis.}, 2020, pp. 241--257.

\bibitem{fedus2022switch}
W.~Fedus, B.~Zoph, and N.~Shazeer, ``Switch transformers: Scaling to trillion
  parameter models with simple and efficient sparsity,'' \emph{Journal of
  Machine Learning Research}, vol.~23, no. 120, pp. 1--39, 2022.

\bibitem{yang2021spatiotemporal}
K.~Yang, D.~Liu, Z.~Chen, F.~Wu, and W.~Li, ``Spatiotemporal generative
  adversarial network-based dynamic texture synthesis for surveillance video
  coding,'' \emph{IEEE Transactions on Circuits and Systems for Video
  Technology}, vol.~32, no.~1, pp. 359--373, 2021.

\bibitem{xie2020spatially}
Z.~Xie, Z.~Zhang, X.~Zhu, G.~Huang, and S.~Lin, ``Spatially adaptive inference
  with stochastic feature sampling and interpolation,'' in \emph{Eur. Conf.
  Comput. Vis.}, 2020, pp. 531--548.

\bibitem{kirillov2020pointrend}
A.~Kirillov, Y.~Wu, K.~He, and R.~Girshick, ``{PointRend}: Image segmentation
  as rendering,'' in \emph{IEEE Conf. Comput. Vis. Pattern Recog.}, 2020, pp.
  9799--9808.

\bibitem{cordonnier2021differentiable}
J.-B. Cordonnier, A.~Mahendran, A.~Dosovitskiy, D.~Weissenborn, J.~Uszkoreit,
  and T.~Unterthiner, ``Differentiable patch selection for image recognition,''
  in \emph{IEEE Conf. Comput. Vis. Pattern Recog.}, 2021, pp. 2351--2360.

\bibitem{yang2020resolution}
L.~Yang, Y.~Han, X.~Chen, S.~Song, J.~Dai, and G.~Huang, ``Resolution adaptive
  networks for efficient inference,'' in \emph{IEEE Conf. Comput. Vis. Pattern
  Recog.}, 2020, pp. 2369--2378.

\bibitem{choy20194d}
C.~Choy, J.~Gwak, and S.~Savarese, ``{4D} spatio-temporal convnets: Minkowski
  convolutional neural networks,'' in \emph{IEEE Conf. Comput. Vis. Pattern
  Recog.}, 2019, pp. 3075--3084.

\bibitem{zhu2021cylindrical}
X.~Zhu, H.~Zhou, T.~Wang, F.~Hong, Y.~Ma, W.~Li, H.~Li, and D.~Lin,
  ``Cylindrical and asymmetrical {3D} convolution networks for lidar
  segmentation,'' in \emph{IEEE Conf. Comput. Vis. Pattern Recog.}, 2021, pp.
  9939--9948.

\bibitem{song2021learning}
Z.~Song, L.~Zhao, and J.~Zhou, ``Learning hybrid semantic affinity for point
  cloud segmentation,'' \emph{IEEE Transactions on Circuits and Systems for
  Video Technology}, vol.~32, no.~7, pp. 4599--4612, 2021.

\bibitem{jiang2020pointgroup}
L.~Jiang, H.~Zhao, S.~Shi, S.~Liu, C.-W. Fu, and J.~Jia, ``{PointGroup}:
  Dual-set point grouping for {3D} instance segmentation,'' in \emph{IEEE Conf.
  Comput. Vis. Pattern Recog.}, 2020, pp. 4867--4876.

\bibitem{li2020multi}
D.~Li, G.~Shi, Y.~Wu, Y.~Yang, and M.~Zhao, ``Multi-scale neighborhood feature
  extraction and aggregation for point cloud segmentation,'' \emph{IEEE
  Transactions on Circuits and Systems for Video Technology}, vol.~31, no.~6,
  pp. 2175--2191, 2020.

\bibitem{shi2020points}
S.~Shi, Z.~Wang, J.~Shi, X.~Wang, and H.~Li, ``From points to parts: {3D}
  object detection from point cloud with part-aware and part-aggregation
  network,'' \emph{IEEE Trans. Pattern Anal. Mach. Intell.}, 2020.

\bibitem{zhao2021transformer3d}
L.~Zhao, J.~Guo, D.~Xu, and L.~Sheng, ``Transformer3d-det: Improving 3d object
  detection by vote refinement,'' \emph{IEEE Transactions on Circuits and
  Systems for Video Technology}, vol.~31, no.~12, pp. 4735--4746, 2021.

\bibitem{deng2021multi}
J.~Deng, W.~Zhou, Y.~Zhang, and H.~Li, ``From multi-view to hollow-3d:
  Hallucinated hollow-3d r-cnn for 3d object detection,'' \emph{IEEE
  Transactions on Circuits and Systems for Video Technology}, vol.~31, no.~12,
  pp. 4722--4734, 2021.

\bibitem{dong2023semantic}
S.~Dong, X.~Kong, X.~Pan, F.~Tang, W.~Li, Y.~Chang, and W.~Dong,
  ``Semantic-context graph network for point-based 3d object detection,''
  \emph{IEEE Transactions on Circuits and Systems for Video Technology}, 2023.

\bibitem{athalye2018obfuscated}
A.~Athalye, N.~Carlini, and D.~Wagner, ``Obfuscated gradients give a false
  sense of security: Circumventing defenses to adversarial examples,'' in
  \emph{Int. Conf. Mach. Learn.}, 2018, pp. 274--283.

\bibitem{glorot2011deep}
X.~Glorot, A.~Bordes, and Y.~Bengio, ``Deep sparse rectifier neural networks,''
  in \emph{Proceedings of the fourteenth international conference on artificial
  intelligence and statistics}, 2011, pp. 315--323.

\bibitem{gao2018dynamic}
X.~Gao, Y.~Zhao, {\L}.~Dudziak, R.~Mullins, and C.-z. Xu, ``Dynamic channel
  pruning: Feature boosting and suppression,'' in \emph{Int. Conf. Learn.
  Represent.}, 2018.

\bibitem{mullapudi2018hydranets}
R.~T. Mullapudi, W.~R. Mark, N.~Shazeer, and K.~Fatahalian, ``{HydraNets}:
  Specialized dynamic architectures for efficient inference,'' in \emph{IEEE
  Conf. Comput. Vis. Pattern Recog.}, 2018, pp. 8080--8089.

\bibitem{qi2017pointnet}
C.~R. Qi, H.~Su, K.~Mo, and L.~J. Guibas, ``{PointNet}: Deep learning on point
  sets for {3D} classification and segmentation,'' in \emph{IEEE Conf. Comput.
  Vis. Pattern Recog.}, 2017, pp. 652--660.

\bibitem{qi2017pointnet++}
C.~R. Qi, L.~Yi, H.~Su, and L.~J. Guibas, ``{PointNet++}: Deep hierarchical
  feature learning on point sets in a metric space,'' in \emph{Adv. Neural
  Inform. Process. Syst.}, 2017, pp. 5105--5114.

\bibitem{wang2019dynamic}
Y.~Wang, Y.~Sun, Z.~Liu, S.~E. Sarma, M.~M. Bronstein, and J.~M. Solomon,
  ``Dynamic graph {CNN} for learning on point clouds,'' \emph{ACM Trans.
  Graph.}, vol.~38, no.~5, pp. 1--12, 2019.

\bibitem{liu2019relation}
Y.~Liu, B.~Fan, S.~Xiang, and C.~Pan, ``Relation-shape convolutional neural
  network for point cloud analysis,'' in \emph{IEEE Conf. Comput. Vis. Pattern
  Recog.}, 2019, pp. 8895--8904.

\bibitem{wu2019pointconv}
W.~Wu, Z.~Qi, and L.~Fuxin, ``{PointConv}: Deep convolutional networks on {3D}
  point clouds,'' in \emph{IEEE Conf. Comput. Vis. Pattern Recog.}, 2019, pp.
  9621--9630.

\bibitem{thomas2019kpconv}
H.~Thomas, C.~R. Qi, J.-E. Deschaud, B.~Marcotegui, F.~Goulette, and L.~J.
  Guibas, ``{KPConv}: Flexible and deformable convolution for point clouds,''
  in \emph{Int. Conf. Comput. Vis.}, 2019, pp. 6411--6420.

\bibitem{deng2021vector}
C.~Deng, O.~Litany, Y.~Duan, A.~Poulenard, A.~Tagliasacchi, and L.~J. Guibas,
  ``Vector neurons: A general framework for {SO(3)}-equivariant networks,'' in
  \emph{Int. Conf. Comput. Vis.}, 2021, pp. 12\,200--12\,209.

\bibitem{tu2020physically}
J.~Tu, M.~Ren, S.~Manivasagam, M.~Liang, B.~Yang, R.~Du, F.~Cheng, and
  R.~Urtasun, ``Physically realizable adversarial examples for lidar object
  detection,'' in \emph{IEEE Conf. Comput. Vis. Pattern Recog.}, 2020, pp.
  13\,716--13\,725.

\bibitem{qian2020end}
R.~Qian, D.~Garg, Y.~Wang, Y.~You, S.~Belongie, B.~Hariharan, M.~Campbell,
  K.~Q. Weinberger, and W.-L. Chao, ``End-to-end pseudo-lidar for image-based
  {3D} object detection,'' in \emph{IEEE Conf. Comput. Vis. Pattern Recog.},
  2020, pp. 5881--5890.

\bibitem{croce2020reliable}
F.~Croce and M.~Hein, ``Reliable evaluation of adversarial robustness with an
  ensemble of diverse parameter-free attacks,'' in \emph{Int. Conf. Mach.
  Learn.}\hskip 1em plus 0.5em minus 0.4em\relax PMLR, 2020, pp. 2206--2216.

\bibitem{almahairi2016dynamic}
A.~Almahairi, N.~Ballas, T.~Cooijmans, Y.~Zheng, H.~Larochelle, and
  A.~Courville, ``Dynamic capacity networks,'' in \emph{Int. Conf. Mach.
  Learn.}, 2016, pp. 2549--2558.

\bibitem{wu20153d}
Z.~Wu, S.~Song, A.~Khosla, F.~Yu, L.~Zhang, X.~Tang, and J.~Xiao, ``{3D}
  shapenets: A deep representation for volumetric shapes,'' in \emph{IEEE Conf.
  Comput. Vis. Pattern Recog.}, 2015, pp. 1912--1920.

\bibitem{tsai2020robust}
T.~Tsai, K.~Yang, T.-Y. Ho, and Y.~Jin, ``Robust adversarial objects against
  deep learning models,'' in \emph{AAAI}, 2020, pp. 954--962.

\end{thebibliography}

%

\begin{IEEEbiography}[{\includegraphics[width=1in,height=1.25in,clip,keepaspectratio]{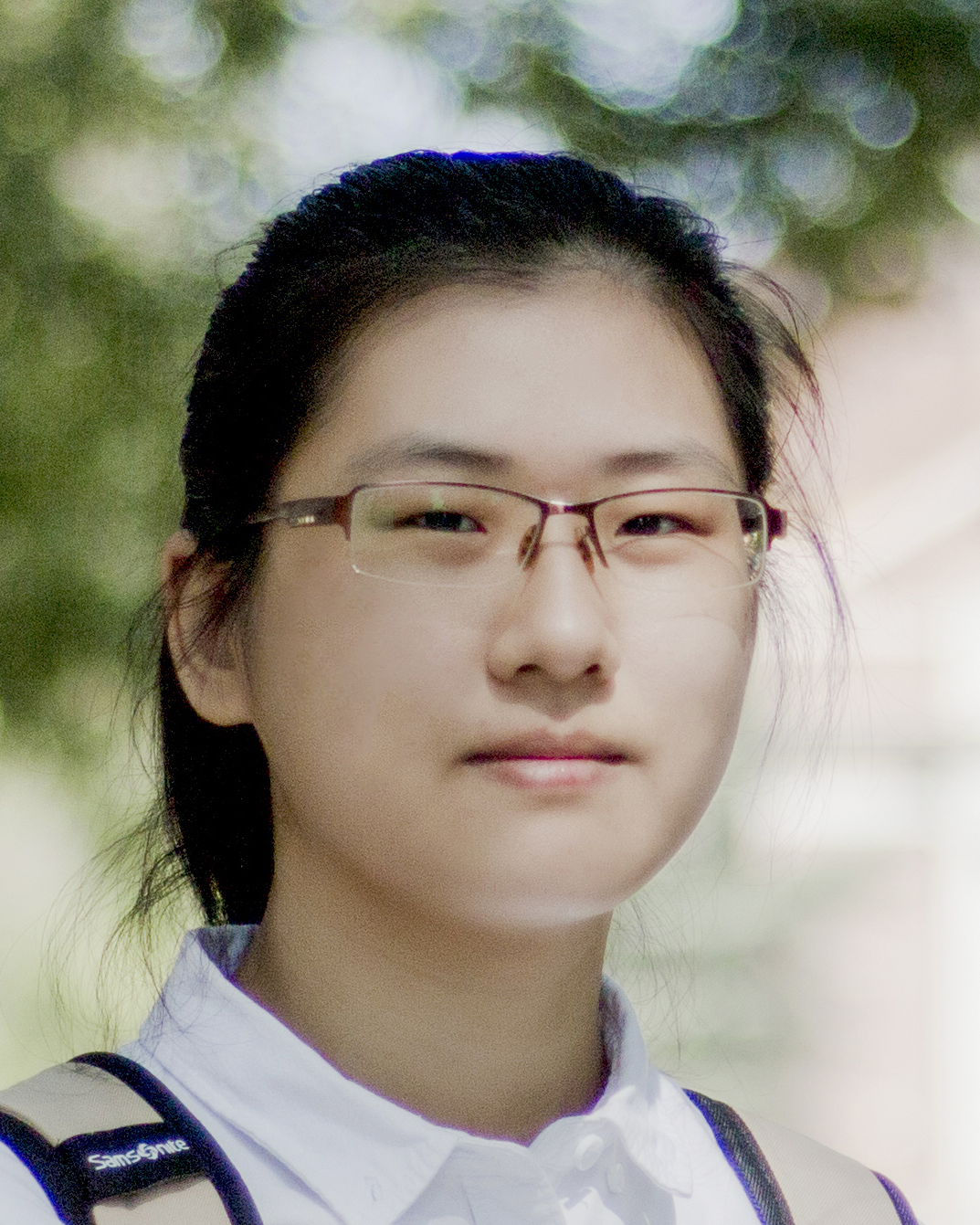}}]{An Tao}
	(Graduate Student Member, IEEE) received the B.Eng. degree from the School of Information Science and Engineering, Southeast University, Nanjing, China, in 2019. She is currently pursuing the Ph.D. degree in the Department of Automation, Tsinghua University, Beijing, China. Her current research interest is 3D vision. She has authored two scientific papers in \textsc{IEEE Transactions on Image Processing} and \textsc{IEEE Transactions on Circuits and Systems for Video Technology}. She has served as a reviewer for several conferences, e.g., CVPR, ICCV, ECCV, ICME, and 3DV.
\end{IEEEbiography}

\vspace{5pt}
\begin{IEEEbiography}[{\includegraphics[width=1in,height=1.25in,clip,keepaspectratio]{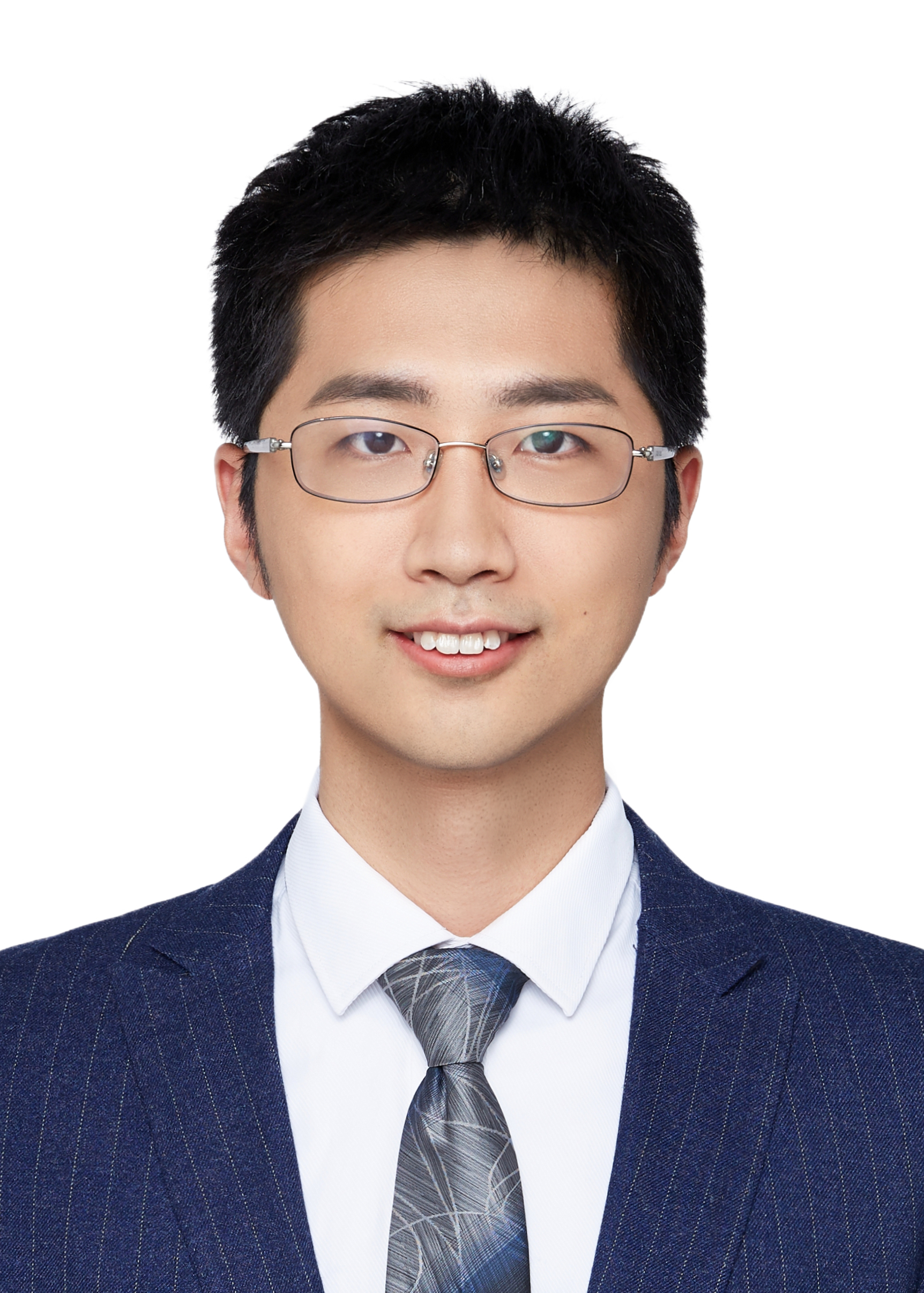}}]{Yueqi Duan}
	(Member, IEEE) received the B.S. and Ph.D. degrees from the Department of Automation, Tsinghua University, in 2014 and 2019, respectively. From 2019 to 2021, he worked as a Postdoctoral Researcher with the Computer Science Department, Stanford University. He is currently an Assistant Professor with the Department of Electronic Engineering, Tsinghua University. He has published more than 20 scientific articles in the top journals and conferences, including \textsc{IEEE Transactions on Pattern Analysis and Machine Intelligence}, \textsc{IEEE Transactions on Image Processing}, CVPR, ICCV and ECCV. His research interests include computer vision and pattern recognition. He served as the Publication Chair of FG, the Area Chair of ICME, and a Regular Reviewer for a number of journals and conferences, e.g., TPAMI, IJCV, TIP, CVPR, ICCV, ECCV, ICML, NeurIPS, and SIGGRAPH. He was awarded the Excellent Doctoral Dissertation of Chinese Association for Artificial Intelligence (CAAI) in 2020.
\end{IEEEbiography}

\vspace{5pt}
\begin{IEEEbiography}[{\includegraphics[width=1in,height=1.25in,clip,keepaspectratio]{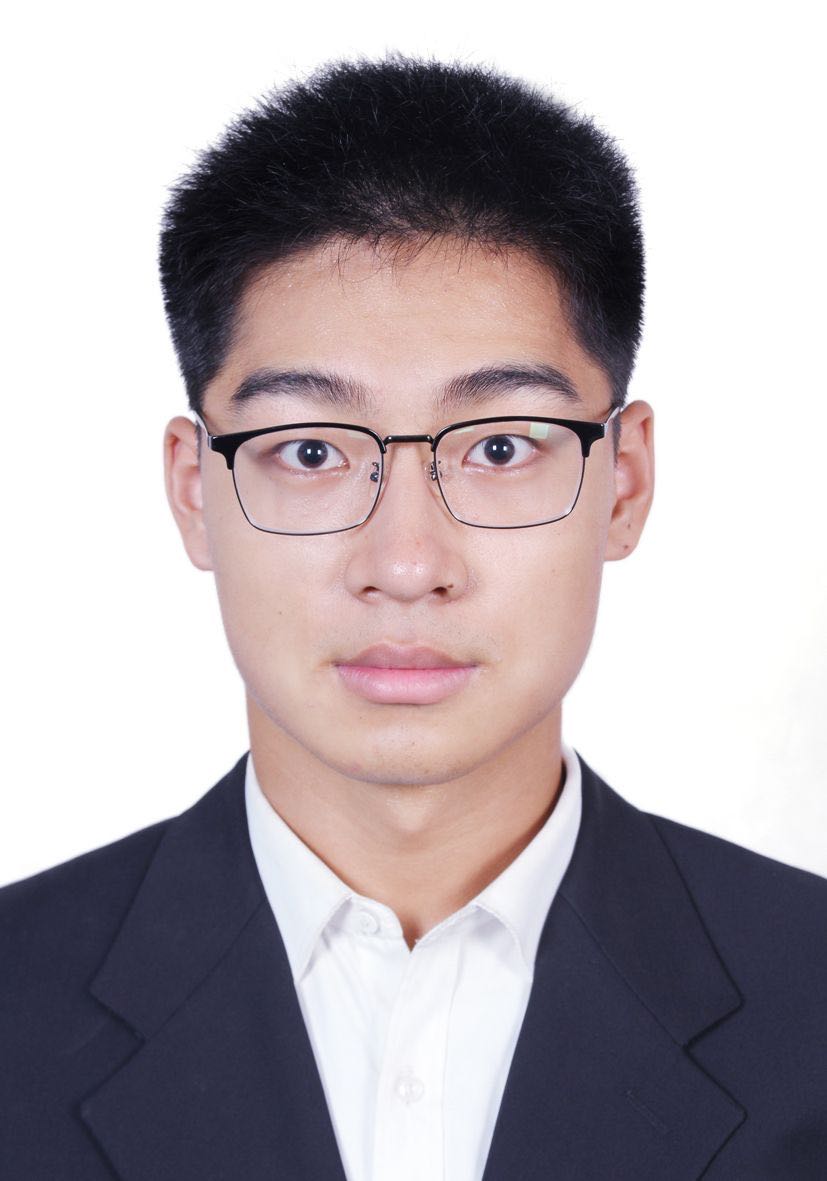}}]{Yingqi Wang}
	is an undergraduate student majoring in Creative Design and Intelligent Engineering at Tsinghua University. His research interests include 3D point cloud processing and AIGC. 
\end{IEEEbiography}

\vspace{5pt}
\begin{IEEEbiography}[{\includegraphics[width=1in,height=1.25in,clip,keepaspectratio]{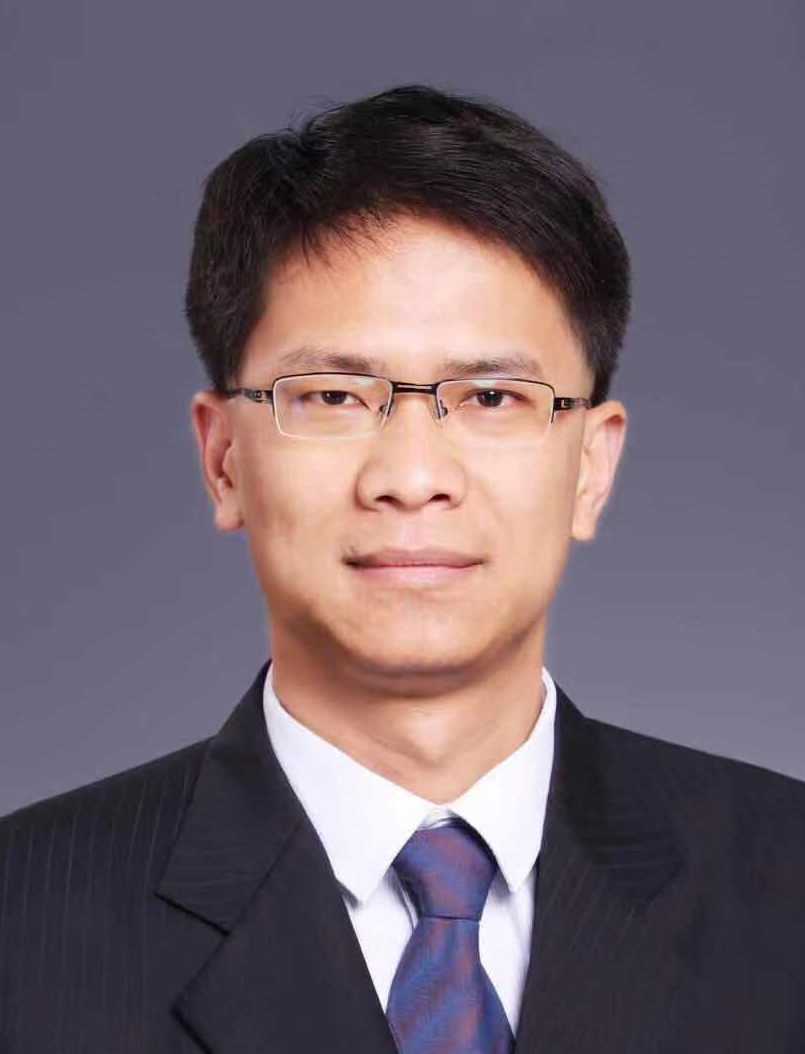}}]{Jiwen Lu}
	(Fellow, IEEE) received the BEng degree in mechanical engineering and the MEng degree in electrical engineering from the Xi’an University of Technology, Xi'an, China, in 2003 and 2006, respectively, and the PhD degree in electrical engineering from Nanyang Technological University, Singapore, in 2012. He is currently an associate professor with the Department of Automation, Tsinghua University, Beijing, China. His current research interests include computer vision and pattern recognition, where he has authored/co-authored more than 160 scientific papers in \textsc{IEEE Transactions on Pattern Analysis and Machine Intelligence}, International Journal of Computer Vision, CVPR, ICCV, and ECCV. He serves as the co-editor-of-chief for Pattern Recognition Letters, an associate editor for \textsc{IEEE Transactions on Image Processing}, \textsc{IEEE Transactions on Circuits and Systems for Video Technology}, \textsc{IEEE Transactions on Biometrics, Behavior, Identity Science, and Pattern Recognition}. He also serves as the program co-chair of FG'2023, VCIP'2022, AVSS'2021, and ICME'2020. He is a recipient of the National Outstanding Youth Foundation of China Award, and an IAPR fellow.
	
\end{IEEEbiography}\vfill

\vspace{5pt}
\begin{IEEEbiography}[{\includegraphics[width=1in,height=1.25in,trim=100 500 100 0,clip,keepaspectratio]{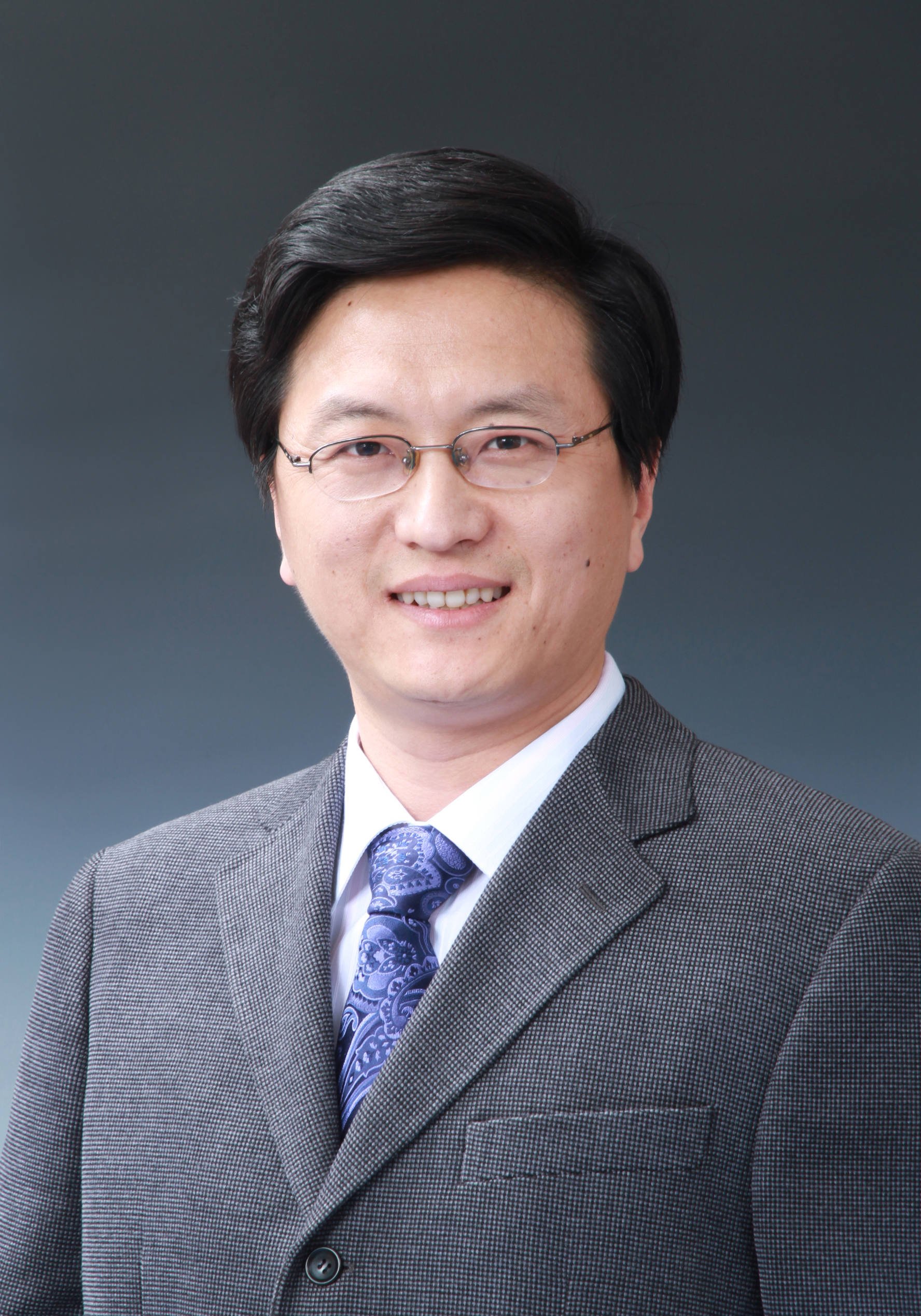}}]{Jie Zhou}
	(Senior Member, IEEE) received the B.S. and M.S. degrees from the Department of Mathematics, Nankai University, Tianjin, China, in 1990 and 1992, respectively, and the Ph.D. degree from the Institute of Pattern Recognition and Artificial Intelligence, Huazhong University of Science and Technology (HUST), Wuhan, China, in 1995. From 1995 to 1997, he was a Postdoctoral Fellow with the Department of Automation, Tsinghua University, Beijing, China, where he has been a Full Professor since 2003. In recent years, he has authored more than 300 papers in peer-reviewed journals and conferences. Among them, more than 100 papers have been published in top journals and conferences such as \textsc{IEEE Transactions on Pattern Analysis and Machine Intelligence}, CVPR, and ICCV. His research interests include pattern recognition, computer vision, and image processing. He is a fellow of IAPR. He received the National Outstanding Youth Foundation of China Award in 2012. He is an Associate Editor of \textsc{IEEE Transactions on Pattern Analysis and Machine Intelligence}.
\end{IEEEbiography}


\vfill
\newpage

\section*{\textbf{Appendix}}
\section{Attack Details on 3D Sparse Convolution Network}
\subsection{Relation Function}
In the main text, we present a sigmoid-like function for occupancy calculation. Two existing functions can also be adopted in this situation, bilinear interpolation~\cite{tu2020physically} and radial basis function~\cite{qian2020end}. The formula of bilinear interpolation (BI) is as follows:
\begin{equation}
r_{\rm BI}(\bm{x}, \bm{x}^{\rm pt}_{n}) = \prod_{i\in\{0,1,2\}}(1-d(\bm{x}, \bm{x}^{\rm pt}_{n})[i])
\end{equation}
where $d(\bm{x}, \bm{x}^{\rm pt}_{n})$ is outputs a distance vector between the voxel center $\bm{x}$ and the point $\bm{x}^{\rm pt}_{n}$.
The formula of radial basis function (RBF) is as follows:
\begin{equation}
r_{\rm RBF}(\bm{x}, \bm{x}^{\rm pt}_{n}) = \frac{1}{\alpha}{\rm exp}(-||d(\bm{x}, \bm{x}^{\rm pt}_{n})||^2)
\end{equation}
where $\alpha$ is a parameter to control the intensity. When the point $\bm{x}^{\rm pt}_{n}$ belongs to the voxel $\bm{x}$, $\alpha=1$, and otherwise $\alpha=26$ (considering 26 neighbors in a $3\times3\times3$ range).

Compared to bilinear interpolation and radial basis function, our sigmoid-like function has two important qualities: 1) the output of the sigmoid-like function is similar to the original hard function in most areas, so the sigmoid-like function can substitute the original function in forwarding propagation; 2) the sigmoid-like function has significant gradient variation near voxel boundaries. Therefore, when a point is located near the voxel boundary, the gradient will become distinctly large to force the point to quickly move in/out of the voxel, instead of staying near the boundary. 

\subsection{End-to-end Attack}
The end-to-end processing pipeline of 3D sparse convolution network includes three parts: 1) voxelization; 2) network processing; 3) devoxelization. In our main text, we focus on part 2 and make 3D sparse convolution differentiable. However, in an overall view, the voxelization and devoxelization process of sparse convolution network also introduces non-differentiable function and thus shatterer gradient propagation. Because the attack target is point cloud data, not voxels, the voxelization and devoxelization also need to be reformulated.

The voxelization process of input $\bm{x}$ is shown in equation (21-23) in the main text. Following the definitions in the main text, we denote the set of features $\mathcal{F}^{\rm pt}=\{\bm{f}^{\rm pt}_n\}^{N}_{n=1}$, and ground-truth labels $\mathcal{Y}^{\rm pt}=\{\bm{y}^{\rm pt}_n\}^{N}_{n=1}$, and the set of features $\mathcal{F}=\{\bm{f}_m\}^{M}_{m=1}$, and ground-truth labels $\mathcal{Y}=\{\bm{y}_m\}^{M}_{m=1}$. We present the voxelization of features and labels:
\begin{equation}
\label{vol}
\{(\bm{f}_m, \bm{y}_m)\}_{m=1}^{M}=\{(\bm{f}^{\rm pt}_n, \bm{y}^{\rm pt}_n)\}_{n\in\mathcal{I}}.
\end{equation}

The 3D sparse convolution network can be defined as:
\begin{equation}
\label{net}
\{\hat{\bm{y}}_m\}_{m=1}^{M} = {\rm SCN}(\{(\bm{x}_m, \bm{f}_m)\}_{m=1}^{M}, \bm{\theta}),
\end{equation}
where $\{\hat{\bm{y}}_m\}_{m=1}^{M}\in \mathbb{R}^{M\times Q}$ is the network output scores for $M$ voxels in $Q$ classes, and $\bm{\theta}$ represents the network parameters. 

Given the network output scores of voxels, the devoxelization can be presented:
\begin{equation}
\label{devol}
\{\hat{\bm{y}}^{\rm pt}_n\}_{n=1}^{N}=\{\hat{\bm{y}}_m|\bm{x}_m=\tilde{\bm{x}}_n\}_{n=1}^{N}
\end{equation}
where $\tilde{\bm{x}}_n$ denotes the floored point clouds and is defined in (21). 

\textbf{Reformulation of Voxellization.} 
Because the normal voxelization process is non-differentiable and dynamics-unaware, we then follow the strategy in leaded gradient formulation to use a differentiable function to interpolate the voxel features with point coordinates and extend the $M$ input voxels into $M'$ voxels. We use $\mathcal{X'}$ to denote the $M'$ voxel coordinates with extended voxels. 
The existence of these $M'$ input voxel features is controlled by the occupancy function. For a voxel $\bm{x}_m\in\mathcal{X'}$, the input voxel feature $\bm{f}_m$ is derived as: 
\begin{gather}
\label{vol_new}
{\tilde {\bm f}}_m = \frac{\sum_{\bm{x}^{\rm pt}_{n}\in \mathcal{N}(\bm{x}_m, \mathcal{X}^{\rm pt})}\hat{r}(\bm{x}_m, \bm{x}^{\rm pt}_{n}){\bm f}_n^{\rm pt}}{\sum_{\bm{x}^{\rm pt}_{n}\in \mathcal{N}(\bm{x}_m, \mathcal{X}^{\rm pt})}\hat{r}(\bm{x}_m, \bm{x}^{\rm pt}_{n})}\\
\label{vol_new2}
{\bm f}_m = \hat{o}(\bm{x}_m,\mathcal{X}^{\rm pt}){\tilde {\bm f}}_m.
\end{gather}
where $\hat{r}(\bm{x}_m, \bm{x}^{\rm pt}_{n})$ and $\hat{o}(\bm{x}_m,\mathcal{X}^{\rm pt})$ are defined in (30) and (31).

\textbf{Reformulation of Devoxelization.} Symmetrical to the voxelization process, we conduct a devoxelization process on the network voxel outputs by interpolating the label scores from $M'$ sparse voxels into $N$ points. Because the $M'$ outputs of the network are already multiplied with occupancy values. we first divide the outputs by their occupancy values to recover the natural values. Then, we interpolate the label scores from $M'$ sparse voxels into $N$ points. The label score $\hat{\bm{y}}^{\rm pt}_n$ for point $\bm{x}^{\rm pt}_n\in \mathcal{X}^{\rm pt}$ is derived as follows: 
\begin{gather}
\label{devol_new}
\tilde{\bm{y}}_m = \frac{\hat{\bm{y}}_m}{\hat{o}(\bm{x}_m,\mathcal{X}^{\rm pt})}\\
\label{devol2}
\hat{\bm{y}}^{\rm pt}_n = \frac{\sum_{\bm{x}_m\in \mathcal{M}(\bm{x}^{\rm pt}_n, \mathcal{X'})}\hat{r}(\bm{x}_m, \bm{x}^{\rm pt}_n)\tilde{\bm{y}}_m}{\sum_{\bm{x}_m\in \mathcal{M}(\bm{x}^{\rm pt}_n, \mathcal{X'})}\hat{r}(\bm{x}_m, \bm{x}^{\rm pt}_n)},
\end{gather}
where $\hat{\bm{y}}_m$ is the output label score for voxel $\bm{x}_m\in\mathcal{X'}$, and $\mathcal{M}(\bm{x}^{\rm pt}_{n}, \mathcal{X'})$ contains the coordinates of a set of voxels that the point $\bm{x}^{\rm pt}_{n}$ possibly exists after one attack step. Therefore, we can conduct the attack point-to-point in an end-to-end way.


\section{More Experiments}
\subsection{Attack Proccess in 2D Images}
\begin{figure*}
	\centering
	\subfigure{
		\begin{minipage}{0.45\linewidth}
			\centering
			\includegraphics[width=1\textwidth, trim=0 0 0 0,clip]{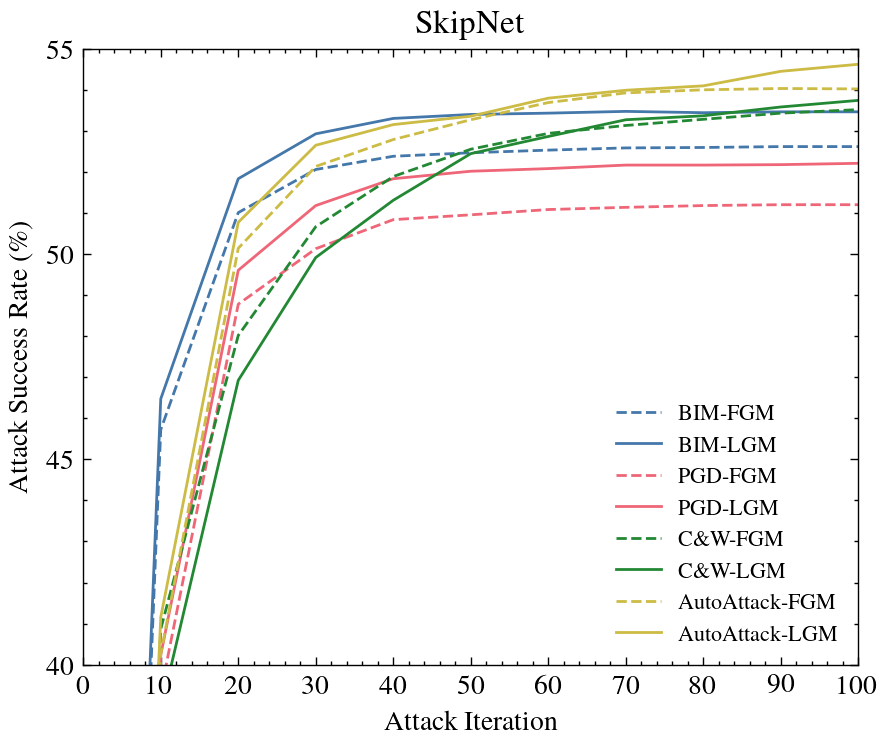}
		\end{minipage}
	}
	\subfigure{
		\begin{minipage}{0.45\linewidth}
			\centering
			\includegraphics[width=1\textwidth, trim=0 0 0 0,clip]{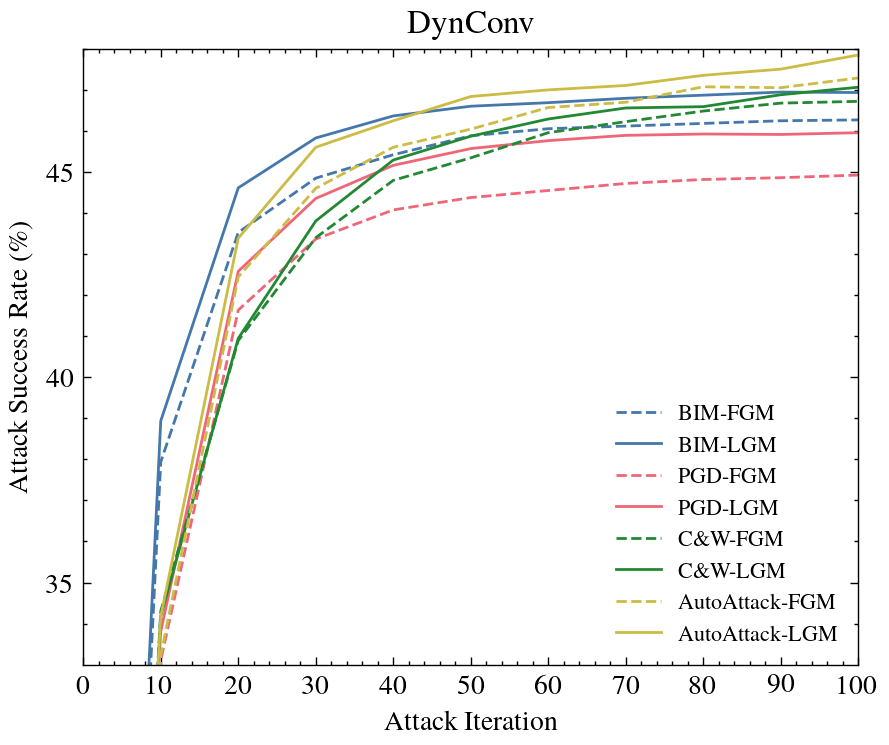}
		\end{minipage}
	}
	\caption{Adversarial attack results of SkipNet (ResNet-110) and DynConv (ResNet-32) on the CIFAR-10 dataset. The valid pixel rate is 5\% and $\epsilon=8$.}
	\label{fig}
\end{figure*}

In Fig.~\ref{fig}, we show the attack success rate in the process of attack of SkipNet and DynConv. The experiments are conducted in the CIFAR-10 dataset with a valid pixel rate 5\% and $\epsilon=8$. We find in almost all cases the superiority of our LGM is apparent at the beginning of the attack. 

\begin{table*}
	\footnotesize
	\caption{Sparse voxel number and GFLOPs during attacks on the ScanNet validation set in different $\epsilon$.}
	\vspace{-3pt}
	{\footnotesize 
		\begin{center}
			\begin{tabular}{l|c|cc|cc}
				\toprule
				\multirow{2}{*}{Method} & ~~~~~\multirow{2}{*}{$\epsilon$}~~~~~ & \multicolumn{2}{c|}{Sparse Voxel Num.} & \multicolumn{2}{c}{GFLOPs}  \\
				& & FGM & LGM & FGM & LGM \\
				\midrule
				Point-BIM~\cite{kurakin2016adversarial}~~~~ & 0.005 & 109.7643K & 117.0876K & 206.0266 & 343.9244 \\
				Point-BIM~\cite{kurakin2016adversarial} & 0.01 & 110.0546K & 354.8085K & 217.1055 & 782.1737 \\
				Point-BIM~\cite{kurakin2016adversarial} & 0.02 & 113.4829K & 406.7601K & 235.6582 & 910.6529 \\
				Point-BIM~\cite{kurakin2016adversarial} & 0.05 & 119.0037K & 475.4461K & 273.0027 & 1075.9983\\
				\bottomrule
			\end{tabular}
		\end{center}
	}
	\label{3d_cost}
\end{table*}

\begin{table*}
	\footnotesize
	\caption{Adversarial attack results (\%) on the ScanNet validation set in different $\lambda$ and $\epsilon$.}
	\vspace{-3pt}
	{\footnotesize 
		\begin{center}
			\begin{tabular}{l|c|cccccccc}
				\toprule
				\multirow{2}{*}{Method} & ~~~~~\multirow{2}{*}{$\epsilon$}~~~~~ & \multicolumn{8}{c}{MinkowskiNet~\cite{choy20194d} (ScanNet) LGM mIoU\textdownarrow} \\
				& & $\lambda=5$ & $\lambda=10$ & $\lambda=15$ & $\lambda=20$ & $\lambda=25$ & $\lambda=30$ & $\lambda=35$ & $\lambda=40$ \\
				\midrule
				Point-BIM~\cite{kurakin2016adversarial}~~~~ & 0.005 & 46.14 & 30.04 & 26.14 & 25.09 & \textbf{24.35} & 25.45 & 24.46 & 25.32 \\
				Point-BIM~\cite{kurakin2016adversarial} & 0.01 & 17.07 & \textbf{10.87} & 12.28 & 14.75 & 16.94 & 18.79 & 20.63 & 20.8  \\
				Point-BIM~\cite{kurakin2016adversarial} & 0.02 & 12.47 & \textbf{5.09} & 6.03 & 6.67 & 7.03 & 7.78 & 8.51 & 8.30  \\
				Point-BIM~\cite{kurakin2016adversarial} & 0.05 & 9.18 & 3.17 & \textbf{2.78} & 3.04 & 3.33 & 3.56 & 3.74 & 3.80\\
				\bottomrule
			\end{tabular}
		\end{center}
	}
	\label{3d_lamb}
\end{table*}

\subsection{Computational Costs between FGM and LGM}
In 2D images, Layer Skipping Network~\cite{wang2018skipnet} and 2D Sparse Convolution Network~\cite{verelst2020dynamic} selectively activate layers/convolutions in their original static network, e.g. ResNet. Therefore, their computation costs are less than their original intact static network. In our dynamics-aware attack, we allow all input-adaptive computation units to activate to let the gradient backward path aware of potential network structure change. As a result, the computational costs of our method are the same as attacking an intact static network, which is illustrated in Figs. 3 and 4. In 3D point clouds, counting computational costs is meaningful, since the sparse convolution operation depends on the point cloud locations. To let the gradient backward dynamics-aware, we need to extra activate some sparse convolution operations, which is illustrated in Fig. 5. The computational costs are thus bigger than dynamics-unaware baseline method. We present the number of sparse voxels and GFLOPs during attacks on the ScanNet dataset in Table~\ref{3d_cost}. We find bigger $\epsilon$ causes larger computational costs. Combining Table~\ref{3d_cost} with Table IV in our manuscript, in $\epsilon=0.05$ the performance of attack is good enough, and the costs of LGM are only around four times of FGM. Theoretically, $3\times3\times3-1=26$ convolutions are available to activate around an existing sparse voxel (Generalized 3D Sparse Conv in Fig. 5), but to save computations we only activate convolutions within step size (Dynamics-aware 3D Sparse Conv in Fig. 5). Therefore, the extra computational costs of LGM in Table~\ref{3d_cost} is tolerable.

\subsection{Impact of Hyperparameter Lambda}
In Table~\ref{3d_lamb} we show the mIoU results of LGM in different $\lambda$ and $\epsilon$. Although the value of $\lambda$ varies in best results in different $\epsilon$, we find the ACC results of different $\lambda$ do not have big variation. All the results of LGM in our manuscript are obtained by linear parameter searching like Table~\ref{3d_lamb}. 

\subsection{Class-specific 3D Semantic Segmentation Results}
In our main text, we present mIoU results on all classes in 3D datasets. We supplement class-specific semantic IoUs on the ScanNet~\cite{dai2017scannet}, S3DIS~\cite{armeni20163d,armeni2017joint}, and SemanticKITTI~\cite{behley2019semantickitti} datasets in Tables~\ref{scannet_class}, \ref{s3dis_class} and~\ref{kitti_class} to show the vulnerability of each class. In the ScanNet dataset, the class \textit{wall}, \textit{bed}, \textit{chair}, \textit{toilet} and \textit{bathtub} also have a certain degree of robustness. This may be explained by the fact that these robust classes have discriminative characteristics in their global shapes or locations, therefore they are difficult to be destroyed through local point perturbation by attacks. 
We find \textit{ceiling}, \textit{chair}, \textit{floor}, \textit{table}, and \textit{wall} are difficult to attack on the S3DIS dataset. These classes are consistent with the robust classes on the ScanNet dataset. For outdoor scenes, we find \textit{car}, \textit{building}, \textit{vegetation} are robust.

\begin{table*}[t]
	\caption{The class-specific semantic IoUs (\%) on the ScanNet validation set in $\epsilon=0.01$ for various methods. }
	\vspace{-3pt}
	{\scriptsize
		\begin{center}
			\begin{tabular}{l|m{0.3cm}<{\centering}m{0.3cm}<{\centering}m{0.3cm}<{\centering}m{0.3cm}<{\centering}m{0.3cm}<{\centering}m{0.3cm}<{\centering}m{0.3cm}<{\centering}m{0.3cm}<{\centering}m{0.3cm}<{\centering}m{0.3cm}<{\centering}m{0.3cm}<{\centering}m{0.3cm}<{\centering}m{0.3cm}<{\centering}m{0.3cm}<{\centering}m{0.3cm}<{\centering}m{0.3cm}<{\centering}m{0.3cm}<{\centering}m{0.3cm}<{\centering}m{0.3cm}<{\centering}m{0.5cm}<{\centering}|m{0.5cm}<{\centering}}
				\toprule
				\raisebox{.mm}{\makecell{Method}} &  \rotatebox{90}{Wall}&\rotatebox{90}{Floor}&\rotatebox{90}{Cab.}&\rotatebox{90}{Bed}&\rotatebox{90}{Chair}&\rotatebox{90}{Sofa}&\rotatebox{90}{Table}&\rotatebox{90}{Door}&\rotatebox{90}{Wind.}&\rotatebox{90}{Bshf.}&\rotatebox{90}{Pic.}&\rotatebox{90}{Cntr.}&\rotatebox{90}{Desk}&\rotatebox{90}{Curt.}&\rotatebox{90}{Fridg.}&\rotatebox{90}{Shwr.}&\rotatebox{90}{Toil.}&\rotatebox{90}{Sink}&\rotatebox{90}{Bath.}&\rotatebox{90}{Ofurn.}& \rotatebox{90}{Mean}\\
				\midrule
				Bf. Attack &  83.29 & 94.81 & 66.02 & 80.98 & 91.22 & 81.65 & 76.27 & 61.41 & 59.12 & 80.73 & 29.51 & 63.30 & 64.26 & 75.88 & 62.06 & 69.15 & 92.09 & 66.75 & 85.93 & 60.01 & 72.22 \\
				\midrule
				Point-BIM~\cite{kurakin2016adversarial} (FGM) & 70.10 & 92.56 & 45.88 & 73.69 & 83.96 & 71.93 & 55.74 & 33.46 & 37.20 & 58.05 & 12.50 & 46.66 & 37.94 & 57.95 & 35.87 & 47.52 & 80.63 & 55.04 & 77.15 & 36.37 & 55.51\\
				Point-BIM~\cite{kurakin2016adversarial} (LGM) &  \textbf{9.52} & \textbf{84.58} & ~\textbf{4.95} & ~\textbf{9.90} & \textbf{26.60} & \textbf{9.98} & \textbf{11.93} & ~\textbf{2.47} & ~\textbf{2.57} & ~\textbf{0.91} & ~\textbf{0.94} & ~\textbf{0.44} & ~\textbf{3.82} & ~\textbf{1.45} & ~\textbf{0.00} & ~\textbf{3.56} & \textbf{15.42} & ~\textbf{5.15} & \textbf{21.80} & \textbf{1.39} & \textbf{10.87}\\
				\bottomrule
			\end{tabular}
		\end{center}
	}
	\label{scannet_class}
\end{table*}

\begin{table*}[t]
	\caption{The class-specific semantic IoUs (\%) on the S3DIS area 5 in budget $\epsilon=0.01~{\rm m}$. }
	\vspace{-3pt}
	{\scriptsize
		\begin{center}
			\begin{tabular}{l|m{0.3cm}<{\centering}m{0.3cm}<{\centering}m{0.3cm}<{\centering}m{0.3cm}<{\centering}m{0.3cm}<{\centering}m{0.3cm}<{\centering}m{0.3cm}<{\centering}m{0.3cm}<{\centering}m{0.3cm}<{\centering}m{0.3cm}<{\centering}m{0.3cm}<{\centering}m{0.3cm}<{\centering}m{0.5cm}<{\centering}|m{0.5cm}<{\centering}}
				\toprule
				\raisebox{.mm}{\makecell{Method}} &  \rotatebox{90}{Clutter} & \rotatebox{90}{Beam} & \rotatebox{90}{Board} & \rotatebox{90}{Bookc.} & \rotatebox{90}{Ceil.} & \rotatebox{90}{Chair} & \rotatebox{90}{Colu.} & \rotatebox{90}{Door} & \rotatebox{90}{Floor} & \rotatebox{90}{Sofa} & \rotatebox{90}{Table} & \rotatebox{90}{Wall} & \rotatebox{90}{Wind.} & \rotatebox{90}{Mean}\\
				\midrule
				Bf. Attack& 53.03 & 0.02 & 65.36 & 67.64 & 92.49 & 89.06 & 36.85 & 71.84 & 97.19 & 71.28 & 73.16 & 81.68 & 51.45 & 65.47 \\
				\midrule
				Point-BIM~\cite{kurakin2016adversarial} (FGM) & 44.09 & 0.00 & 43.18 & 59.67 & 89.44 & 77.62 & 27.67 & 45.23 & 92.01 & 34.85 & 64.96 & 66.60 & 35.31 & 52.35\\
				Point-BIM~\cite{kurakin2016adversarial} (LGM) & \textbf{31.16} & 0.00 & \textbf{21.48} & \textbf{43.26} & \textbf{86.11} & \textbf{71.25} & \textbf{10.71} & \textbf{31.93} & \textbf{88.40} & \textbf{16.14} & \textbf{52.63} & \textbf{51.18} & \textbf{11.20} & \textbf{39.65}\\
				\bottomrule
			\end{tabular}
		\end{center}
	}
	\label{s3dis_class}
\end{table*}

\begin{table*}[t]
	\caption{The class-specific semantic IoUs (\%) on the SemanticKITTI validation set in budget $\epsilon=0.02~{\rm m}$. }
	\vspace{-3pt}
	{\scriptsize
		\begin{center}
			\begin{tabular}{l|m{0.3cm}<{\centering}m{0.3cm}<{\centering}m{0.3cm}<{\centering}m{0.3cm}<{\centering}m{0.3cm}<{\centering}m{0.3cm}<{\centering}m{0.3cm}<{\centering}m{0.3cm}<{\centering}m{0.3cm}<{\centering}m{0.3cm}<{\centering}m{0.3cm}<{\centering}m{0.3cm}<{\centering}m{0.3cm}<{\centering}m{0.3cm}<{\centering}m{0.3cm}<{\centering}m{0.3cm}<{\centering}m{0.3cm}<{\centering}m{0.3cm}<{\centering}m{0.5cm}<{\centering}|m{0.5cm}<{\centering}}
				\toprule
				\raisebox{.mm}{\makecell{Method}} &  \rotatebox{90}{Car} & \rotatebox{90}{Bic.} & \rotatebox{90}{Mot.} & \rotatebox{90}{Truck} & \rotatebox{90}{Bus} & \rotatebox{90}{Pers.} & \rotatebox{90}{Biclt.} & \rotatebox{90}{Motlt.} & \rotatebox{90}{Road} & \rotatebox{90}{Park.} & \rotatebox{90}{Sidew.} & \rotatebox{90}{Other.} & \rotatebox{90}{Build.} & \rotatebox{90}{Fence} & \rotatebox{90}{Vege.} & \rotatebox{90}{Trunk} & \rotatebox{90}{Teran.} & \rotatebox{90}{Pole} & \rotatebox{90}{Traffic.} & \rotatebox{90}{Mean}\\
				\midrule
				Bf. Attack &  97.12 & 54.47 & 80.86 & 85.10 & 70.34 & 76.48 & 92.22 & 0.02 & 94.58 & 44.84 & 81.19 & 0.95 & 90.48 & 58.71 & 86.64 & 70.82 & 70.48 & 64.23 & 51.79 & 66.91 \\
				\midrule
				Point-BIM~\cite{kurakin2016adversarial} (FGM) & 59.25 & 12.43 & 8.44 & \textbf{1.32} & 3.54 & 18.72 & 29.14 & 0.00 & 6.70 & \textbf{0.67} & 5.22 & 0.01 & 41.44 & \textbf{1.42} & 46.15 & 21.83 & 8.92 & 20.88 & 18.20 & 16.01\\
				Point-BIM~\cite{kurakin2016adversarial} (LGM) & \textbf{55.84} & \textbf{11.13} & \textbf{6.44} & 1.62 & \textbf{3.14} & \textbf{18.00} & \textbf{27.58} & 0.00 & 6.70 & 0.68 & \textbf{5.14} & 0.01 & \textbf{39.33} & 1.47 & \textbf{44.98} & \textbf{20.50} & \textbf{7.80} & \textbf{20.71} & \textbf{17.22} & \textbf{15.17}\\
				\bottomrule
			\end{tabular}
		\end{center}
	}
	\label{kitti_class}
\end{table*}

\subsection{Object-level 3D Classification.} 
In our main text, we conduct experiments on point cloud scene datasets (ScanNet~\cite{dai2017scannet}, S3DIS~\cite{armeni20163d,armeni2017joint}, and SemanticKITTI~\cite{behley2019semantickitti}). To further validate the effectiveness of our method, we also conduct experiments on an object-level dataset ModelNet40~\cite{wu20153d}.
This dataset contains 12,311 3D CAD models from 40 common object categories in the world. It splits 9,843 objects for the training set and 2,468 for the testing set. We follow the data preparation proposed by Qi~\textit{et al.}~\cite{qi2017pointnet} to uniformly sample 2,048 points from the surface of each object. For adversarial attacks, we perturb point clouds in the testing set.

We adopt MinkowskiNet~\cite{choy20194d} on the ModelNet40 dataset for object classification. Table~\ref{modelnet} shows the attack results in different budgets. In all budgets, our LGM outperforms the baseline method FGM and random perturbation, which demonstrates the effectiveness of our dynamics-aware strategy. Since the point cloud features on the ModelNet40 dataset also include point cloud coordinates, the overall performance drop between LGM and FGM compared to the drop between FGM and random perturbation is similar to the SemanticKITTI dataset. We also observe a distinct performance drop when the budget is close to the voxel size 0.05 m (5 cm). Compared with the state-of-the-art point perturbation methods PointPert~\cite{xiang2019generating} and KNNPert~\cite{tsai2020robust} that do not consider the dynamic network changes of 3D sparse convolution network, our LGM outperforms them in all budgets. As LGM is designed based on fundamental gradients, we can build the ``leaded version'' of existing attack methods to further boost their performance. Experimental results in Table~\ref{modelnet} validate the generalization ability of our LGM.

\begin{figure}[t]
	\begin{center}
		\includegraphics[width=1\linewidth, trim=0 0 0 0,clip]{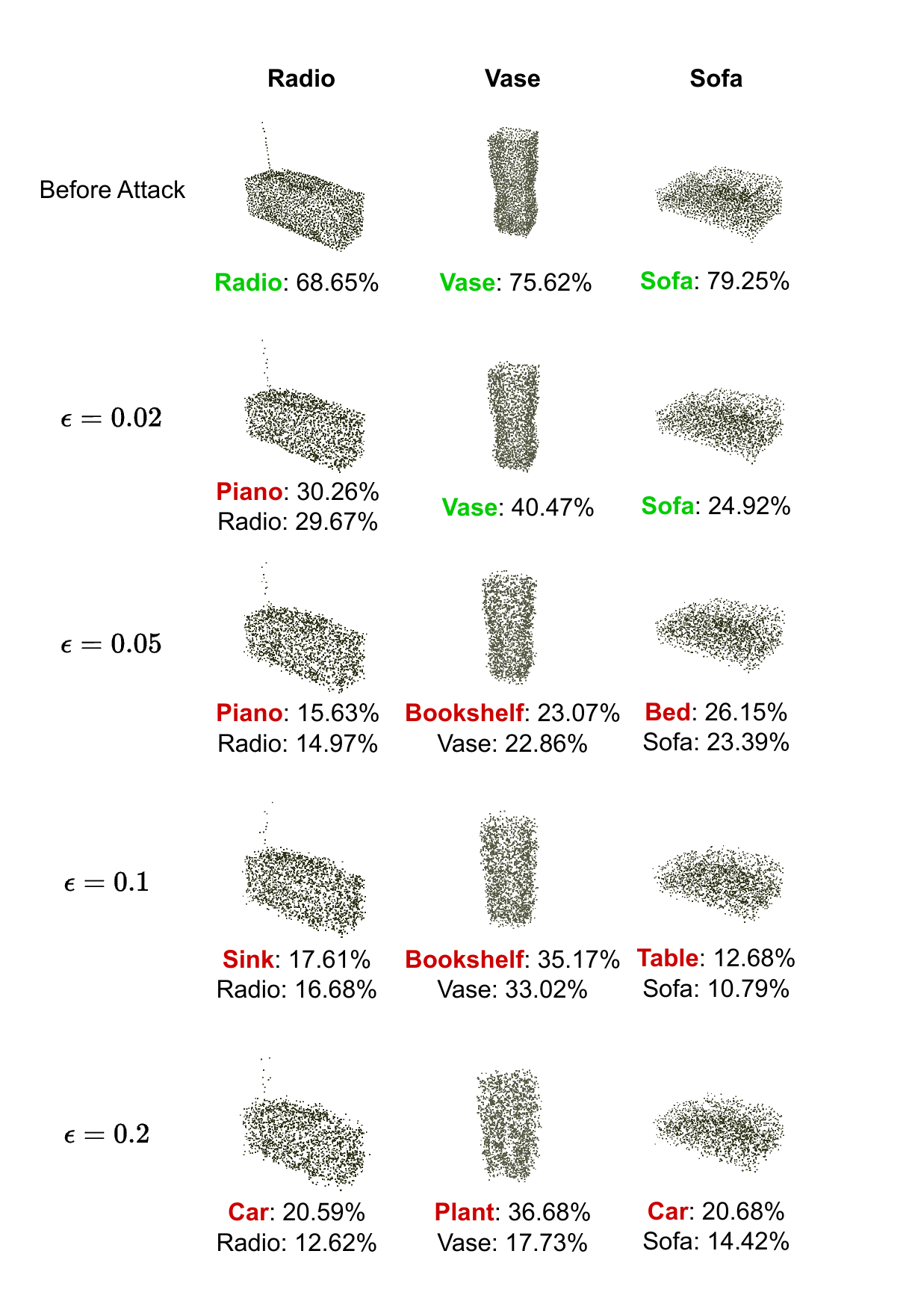}
	\end{center}
	\vspace{-20pt}
	\caption{The qualitative visualization results of LGM on the ModelNet40 testing set in different $\epsilon$ (m).}
	\label{object}
\end{figure}

Figure~\ref{object} shows the visualization results of our LGM in different budgets. We adopt green color to denote the class prediction is right, and red color to denote the class prediction is wrong. We enclose each prediction with a normalized confidence score output from the network. Because the objects are normalized into a unit sphere in the data preprocessing step, the budget unit may not have any reference meaning in the real scene. From the visualized objects, we observed that the perturbation is distinct for all budgets. Compared to the attack performance under some imperceptible budgets on the ScanNet, S3DIS, and SemanticKITTI datasets for semantic segmentation, 3D sparse convolution network is relatively more robust to adversarial attacks on the ModelNet40 dataset for object classification. From the visualization results, we observe that the wrong prediction of each attacked object does to some extent describe the point cloud shape. Therefore, the attack forces the object shape to approach the boundaries between its true class and other classes to fool the victim network. This phenomenon reflects the victim network does learn a certain degree of discriminative information for each class.

\begin{table}[t]
	\footnotesize
	\caption{Point cloud classification accuracy results (\%) on the ModelNet40 testing set in different $\epsilon$ (m) for various methods.}
	\vspace{-3pt}
	{\scriptsize
		\begin{center}
			\begin{tabular}{l|m{1cm}<{\centering}m{1cm}<{\centering}m{1cm}<{\centering}m{1cm}<{\centering}}
				\toprule
				Method & $\epsilon=0.02$ & $\epsilon=0.05$ & $\epsilon=0.1$ & $\epsilon=0.2$\\
				\midrule
				Bf. Attack &\multicolumn{4}{c}{92.38}\\
				\midrule
				Random Perturbation & 85.73 & 79.09 & 41.73 & 6.48\\
				\midrule
				Point-BIM~\cite{kurakin2016adversarial} (FGM) & 83.10 & 64.06 & 25.85 & 3.81\\
				Point-BIM~\cite{kurakin2016adversarial} (LGM) & \textbf{82.46} & \textbf{55.15} & \textbf{19.12} & \textbf{3.12}\\
				\midrule
				PointPert~\cite{xiang2019generating} (FGM) & 84.40 & 72.57 & 35.90 & 5.51\\
				PointPert~\cite{xiang2019generating} (LGM) & \textbf{82.65} & \textbf{55.92} & \textbf{21.47} & \textbf{2.79}\\
				\midrule
				KNNPert~\cite{tsai2020robust} (FGM) & 82.62 & 59.32 & 19.81 & 4.90\\
				KNNPert~\cite{tsai2020robust} (LGM) & \textbf{82.05} & \textbf{48.50} & \textbf{8.95} & \textbf{2.96}\\
				\bottomrule
			\end{tabular}
		\end{center}
	}
	\label{modelnet}
\end{table}

\end{document}